%% file: main.tex
\documentclass[10pt]{article} 
\usepackage[preprint]{tmlr}

\usepackage{amsmath}
\input{math_commands.tex}

\usepackage{url}

\usepackage{graphicx}
\usepackage{booktabs}
\usepackage{multirow}
\usepackage{longtable}
\usepackage{algorithm}
\usepackage{algpseudocode}

\usepackage{hyperref}

\title{Disentangling Steering Vectors}

\author{
    Takeru Hiramatsu$^{1\thanks{
hiramatsu.takeru.64v@st.kyoto-u.ac.jp}}$\hspace{3pt},\quad Kyohei Atarashi$^1$,\quad Koh Takeuchi$^1$ ,\quad Hisashi Kashima$^1$\\
    \\
    $^{1}$\textnormal{\textit{Kyoto University}}\\
    }

\begin{document}

\maketitle

\begin{abstract}
Activation steering has emerged as a lightweight, inference-time approach to control the behavior of Large Language Models (LLMs). However, traditional steering vectors used to intervene in LLMs' activations, such as those derived from the difference-in-means method, tend to entangle multiple semantic and stylistic concepts into a single composite direction, leading to unpredictable steering effects. Our core objective is to disentangle~\footnote{We use the term ``disentanglement'' to refer to separating representations into semantically consistent and interpretable features, rather than to disentanglement in the strict sense of disentangled representation learning.} this composite direction into its constituent concepts. To this end, we propose Steering Vector Dissection, a framework to explicitly isolate individual and semantically consistent features from these composite directions. Specifically, we pair positive and negative activations and take their differences to generate a set of instance-level steering vectors, and train a dedicated Sparse Autoencoder (SAE) directly on them. Quantitative evaluations across two datasets, two models, and two intervention depths show that our method yields a set of semantically consistent basis vectors whose steering effects are mutually distinguishable. Furthermore, we show that this disentanglement enables precise control over model behaviors~\footnote{We used AI tools (e.g., ChatGPT) for proofreading and language editing.}.
\end{abstract}

\section{Introduction}
\label{introduction}
As the capabilities and deployment of LLMs continue to expand, control over their behavior has become an important issue~\citep{ngo2024the}. Previous research highlights that LLMs can exhibit undesirable behaviors, such as sycophancy~\citep{ai_sycophant}. While preference-based alignment methods such as DPO~\citep{rafailov2023direct} steer models away from unintended behaviors, these approaches are inherently black-box approaches. Furthermore, recent studies have shown that the safety alignment established by such methods can be easily undone through lightweight fine-tuning~\citep{yang2024shadow, qi2024finetuning}. To address the limitations of these black-box approaches, representation engineering has emerged as a white-box alternative for monitoring and controlling model behavior~\citep{wehner2025taxonomy, zou2025representationengineeringtopdownapproach}. Rather than altering the model's weights, representation engineering achieves behavioral control by directly intervening in the LLM's internal representations during inference.

One of the simplest and most common interventions is Linear Activation Addition, where a fixed vector---often referred to as a steering vector---is added to the model's activations to inject specific concepts. This method is grounded in the linear representation hypothesis~\citep{park2023the}, which posits that concepts are represented as linear directions within the representation space of LLMs. Various methods have been proposed to calculate these concept-specific directions. For example, Activation Addition~\citep{turner2024steeringlanguagemodelsactivation} and Contrastive Activation Addition~\citep{rimsky-etal-2024-steering} compute the difference in activations between inputs with and without the target concepts. Another approach utilizes linear probes~\citep{li2023inferencetime}, which are classifiers (such as logistic regression) that learn to predict whether activations originate from positive or negative inputs. The learned weights of these classifiers are then used as a directional representation of the concept.

Although representation engineering via a single steering vector has proven effective, it still faces significant challenges. A fundamental limitation is that the previously described methods often fail to isolate specific, individual concepts. For instance, steering a model with a ``happiness'' vector can inadvertently cause it to comply with harmful requests~\citep{zou2025representationengineeringtopdownapproach}. In another case, steering along a ``gender bias awareness'' feature was found to simultaneously exacerbate age bias~\citep{durmus2024steering}. These unintended side effects indicate that a single steering direction often encapsulates multiple entangled concepts, leading to unpredictable downstream behaviors. This raises a fundamental question: can a conventional steering vector be understood as a composite object whose constituent behavioral directions can be isolated and selectively manipulated?

\begin{figure}
 \begin{center}
 \includegraphics[width=1.0\linewidth]{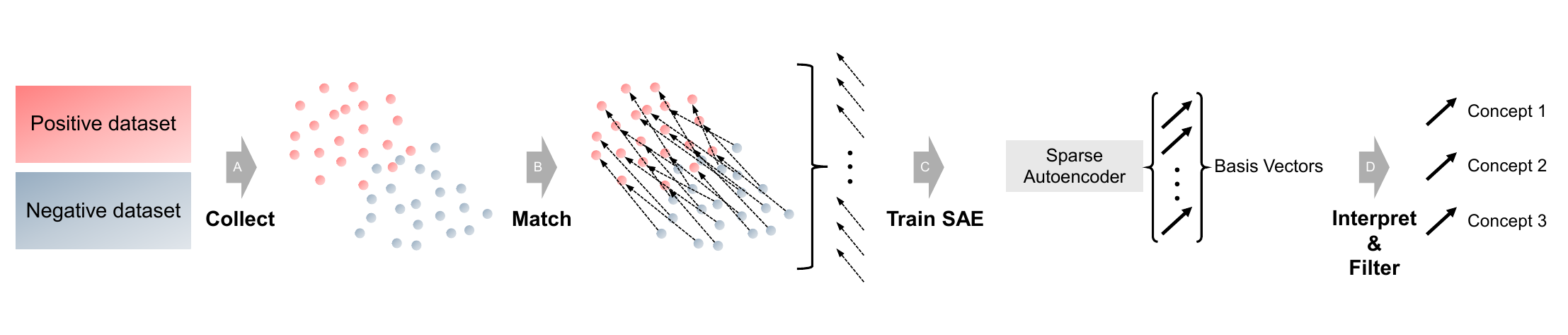}
 \caption{Overview of the proposed Steering Vector Dissection pipeline. (A) Collect model activations using positive and negative datasets. (B) Pair positive and negative activations to compute a set of steering vectors. (C) Train a Sparse Autoencoder (SAE) on the steering vectors, uncovering the basis vectors underlying them. (D) Interpret and filter the resulting basis vectors to isolate interpretable features.}
 \label{fig:pipeline}
 \end{center}
\end{figure}

To address this challenge and explicitly disentangle composite steering directions, we propose Steering Vector Dissection (Figure~\ref{fig:pipeline}). Our approach specifically focuses on vectors derived from activation differences. By pairing the positive and negative activations via nearest-neighbor search and training a dedicated Sparse Autoencoder (SAE) directly on the differences between the paired activations, we extract a set of basis vectors. Through comprehensive analyses of two datasets—one domain-specific and one behavior-related—we demonstrate that the basis vectors obtained via our method capture specific and semantically consistent concepts. An additional experiment shows that our method enables targeted interventions by removing unwanted directions identified via our method from DiffMean vectors. Ultimately, our method represents a step toward precise control over the model's behavior using these disentangled steering vectors.

In summary, the main contributions of this work are twofold:

\begin{itemize}
    \item \textbf{Disentangling Composite Steering Vectors}: We introduce Steering Vector Dissection, a pipeline to disentangle composite steering vectors into semantically consistent basis vectors. We provide quantitative and visual evidence that a traditional composite steering vector contains multiple concepts, and the corresponding basis vectors cover distinct, non-overlapping sub-regions of the concept space induced by the DiffMean vector.
    \item \textbf{Targeted Behavioral Control}: We show that an unsafe DiffMean vector contains rejection-related components as constituents, and that selectively removing them yields targeted changes in model behavior—demonstrating that the extracted directions are behaviorally meaningful, not merely interpretable.
\end{itemize}

\section{Preliminaries}
\label{preliminary}
Our main focus is on disentangling steering vectors. In our proposed framework, we use an SAE for this purpose. Here, we describe activation steering and SAEs.

\subsection{Activation Steering}
\label{pre:activation_steering}
Representation-based methods for controlling model behavior have recently gained significant attention. Unlike black-box methods such as standard fine-tuning, representation-based methods directly intervene in the internal representations of LLMs. Although representation-based fine-tuning methods~\citep{wu2024reft} have proven effective, simpler steering strategies proposed earlier remain widely used. For example, Inference-Time Intervention (ITI)~\citep{li2023inferencetime} works by shifting the outputs of selected attention heads toward a ``truthful'' direction, inducing more truthful behavior in LLMs. 

Methods such as ITI require identifying directions in activation space that represent a target behavior or concept. A widely adopted approach to constructing such directions is to calculate the difference in mean activations between positive and negative text datasets. The resulting vector is commonly referred to as the difference-in-means (DiffMean) vector. Formally, given an LLM with hidden dimension $d$ and a dataset of $N$ prompt-response pairs, let $n_i$ denote the number of response tokens for the $i$-th pair. The activations at a specific layer $l$ for these response tokens are represented as a sequence of vectors $\mathbf{A}_i^l = [\mathbf{a}_1^l, \mathbf{a}_2^l, \dots, \mathbf{a}_{n_i}^l] \in \mathbb{R}^{n_i \times d}$. By collecting all such token-wise activations across the dataset, we construct an aggregated activation set $\mathcal{A}$, containing $\sum_{i=1}^{N} n_i$ vectors. The DiffMean vector is then computed as:

\begin{equation}
\mathbf{v}_{\text{DiffMean}} = \underbrace{\frac{1}{|\mathcal{A}_{\mathrm{pos}}|} \sum_{\mathbf{a}^+ \in \mathcal{A}_{\mathrm{pos}}} \mathbf{a}^+}_{\text{mean of positives}} - \underbrace{\frac{1}{|\mathcal{A}_{\mathrm{neg}}|} \sum_{\mathbf{a}^- \in \mathcal{A}_{\mathrm{neg}}} \mathbf{a}^-}_{\text{mean of negatives}},
\end{equation}

where $\mathcal{A}_{\mathrm{pos}}$ and $\mathcal{A}_{\mathrm{neg}}$ denote the sets of token-wise activations from the positive and negative datasets, respectively. Following prior work~\citep{wu2025axbench}, we compute this global mean across all individual tokens rather than taking the macro-average of sequence-level means from each prompt-response pair.

This simple method has been shown to be capable of steering LLM behavior toward specific concepts such as ``love'' or ``weddings''~\citep{turner2024steeringlanguagemodelsactivation}, specific personas like ``sycophancy'' or ``evil''~\citep{chen2025personavectorsmonitoringcontrolling, rimsky-etal-2024-steering}, distinct emotions~\citep{zou2025representationengineeringtopdownapproach}, specific roles~\citep{poterti-etal-2025-role}, or even enhancing LLMs' instruction-following capabilities~\citep{stolfo2025improving}. However, as described in the introduction, despite their simplicity and effectiveness, these steering vectors are entangled; they contain multiple concepts within a single direction. Consequently, isolating specific concept representations remains a critical challenge in representation engineering~\citep{wehner2025taxonomy}.

\subsection{Sparse Autoencoders}
\label{pre:sae}
Mechanistic interpretability---the study of the internal workings of LLMs---has recently garnered significant attention, largely driven by the success of SAEs~\citep{makhzani2014ksparseautoencoders} in extracting interpretable features from LLM activations~\citep{bricken2023monosemanticity, huben2024sparse, templeton2024scaling}. In neural networks, interpreting internal states in terms of individual neurons is challenging because a single neuron often activates for semantically unrelated inputs, a phenomenon known as \textit{polysemanticity}~\citep{olah2020zoom, bricken2023monosemanticity}. A primary cause of this is \textit{superposition}, where neural networks represent more features than their available dimensions~\citep{elhage2022superposition}. To overcome this bottleneck, SAEs project LLM activations into a higher-dimensional space, disentangling them into interpretable, monosemantic features.

Formally, an SAE is defined as follows:
\begin{align}
\mathbf{f}(\mathbf{x}) &= \sigma(\mathbf{W}_{\mathrm{enc}}\mathbf{x} + \mathbf{b}_{\mathrm{enc}}), \\
\hat{\mathbf{x}} &= \mathbf{W}_{\mathrm{dec}}\mathbf{f}(\mathbf{x}) + \mathbf{b}_{\mathrm{dec}},
\end{align}
where $\sigma$ is an activation function that imposes sparsity. In some architectures, $\sigma$ also enforces non-negativity~\citep{rajamanoharan2024jumpingaheadimprovingreconstruction, bussmann2025learning}.

Beyond raw LLM activations, SAEs have demonstrated the capability to disentangle dense text embeddings~\citep{oneill2024disentanglingdenseembeddingssparse, movva2025sparse}, as well as the differences between activations~\citep{joshi2026sparse} or paired text embeddings~\citep{movva2026whats}. Our research builds directly upon these insights. Because taking the difference in activations is precisely how DiffMean steering vectors are calculated, applying SAEs to these activation differences provides a principled framework for disentangling composite steering vectors into their foundational concepts.

Our novelty does not lie in applying an SAE to activation differences per se. Rather, we study whether conventional steering vectors can be treated as composite objects whose constituent behavioral directions can be identified and selectively edited.

\section{Steering Vector Dissection}
\label{sec:steering_vector_dissection}
Our proposed framework for disentangling steering vectors, termed Steering Vector Dissection, consists of four key steps. The overall methodology builds upon the frameworks established in prior work~\citep{movva2026whats, joshi2026sparse}. Importantly, our method focuses on disentangling steering vectors derived from differences in the output hidden states of the selected transformer layer, and applies steering at the same location during generation.

First, we collect token-wise activations from positive and negative datasets and pair them using nearest-neighbor search to construct a set of steering vectors. Second, we train an SAE on these steering vectors. Owing to the architecture of our SAE, the column vectors of its decoder provide candidate directions underlying the input steering vectors. Third, we interpret the concepts represented by these decoder vectors through steering experiments. Finally, we evaluate and filter the resulting vectors. We describe each step in detail below.

\subsection{Collect and Match Token-wise Activations} 
Our initial objective is to construct a collection of raw steering vectors that capture the semantic difference between the positive and negative datasets. To achieve this, we process prompt-response pairs from both datasets through an instruction-tuned LLM and extract their token-wise activations. Let

\begin{equation}
\mathcal{A}_{\mathrm{pos}} = \{\mathbf{a}^{+}_i \in \mathbb{R}^d\}_{i=1}^{N_+},
\qquad
\mathcal{A}_{\mathrm{neg}} = \{\mathbf{a}^{-}_j \in \mathbb{R}^d\}_{j=1}^{N_-}
\end{equation}

denote the resulting positive and negative activations, respectively.

We then align the positive and negative activations using batch-wise nearest-neighbor search. For each batch $b$, let $\mathcal{A}^{b}_{\mathrm{pos}}$ and $\mathcal{A}^{b}_{\mathrm{neg}}$ denote the positive and negative activations contained in the batch. For each positive activation $\mathbf{a}^{+}_i \in \mathcal{A}^{b}_{\mathrm{pos}}$, we identify its nearest negative activation under cosine distance:

\begin{equation}
j^{*}(i) = \argmin_{j:\mathbf{a}^{-}_j \in \mathcal{A}^{b}_{\mathrm{neg}}} 1 -
\frac{\langle \mathbf{a}^{+}_{i}, \mathbf{a}^{-}_j \rangle}
{\|\mathbf{a}^{+}_i\|_2 \|\mathbf{a}^{-}_j\|_2},
\end{equation}

where $\langle \cdot, \cdot \rangle$ denotes the inner product and $\|\cdot\|_2$ denotes the $\ell_2$-norm.

Pairings are not constrained to be one-to-one; consequently, the same negative activation may be paired with multiple positive activations. Performing this search batch-wise prevents a small number of negative activations from dominating the pairings.

Intuitively, each activation contains information about both the corresponding token itself and its context. Aligning the activations therefore makes it possible to isolate the contextual difference by subtraction.

Finally, for each matched pair, we define an instance-level steering vector as

\begin{equation}
\mathbf{v}_i
=
\mathbf{a}^{+}_i
-
\mathbf{a}^{-}_{j^{*}(i)}.
\end{equation}

The resulting collection
\begin{equation}
\mathcal{V}
=
\{\mathbf{v}_i\}_{i=1}^{N_+}
\end{equation}

serves as the foundational set of raw steering vectors used in the subsequent stages of our method.

\subsection{Train an SAE} 
We adopt the BatchTopK SAE architecture from prior work~\citep{bussmann2024batchtopk}, and modify it with the following key features:
\begin{itemize}
    \item \textbf{Low-dimensional latent space:} Our SAE utilizes a very small number of latent dimensions compared to SAEs trained on raw activations from general texts. As described in prior work~\citep{movva2025sparse, movva2026whats}, when the expected number of features is small, an excessively large latent space could become redundant. Because our input data consists of steering vectors derived from a specific dataset, the underlying generative features are expected to be few in number. 
    
    \item \textbf{Bias-free decoder:} The decoder of our SAE does not include a bias term. This ensures that the reconstructed steering vectors are strictly a linear combination of the column vectors in the decoder's weight matrix. Without a bias term, these column vectors function as the basis vectors that generate the steering vectors. Formally, our SAE is defined as follows:
    \begin{align}
    \mathbf{f}(\mathbf{x}) &= \sigma(\mathrm{ReLU}(\mathbf{W}_{\mathrm{enc}}\mathbf{x} + \mathbf{b}_{\mathrm{enc}})), \\
    \hat{\mathbf{x}} &= \mathbf{W}_{\mathrm{dec}}\mathbf{f}(\mathbf{x}).
    \end{align}
    Here, during training, $\sigma$ is the BatchTopK function, which sets all activation values to zero that are not among the top $Bk$ in the batch, where $B$ is the batch size and $k$ is the target average number of active latents per sample. During inference, following the original BatchTopK SAE, $\sigma$ is the JumpReLU function, which is defined as follows:
    \begin{equation}
    \mathrm{JumpReLU}_{\theta}(z) = z H(z-\theta),
    \end{equation}
    where $H$ is the Heaviside step function and $\theta$ is a global threshold estimated from training batches. The training objective is the same as in the BatchTopK SAE~\citep{bussmann2024batchtopk}.

    \item \textbf{No data centering:} Typically, SAEs are trained on centered inputs. However, because our primary objective is to capture the low-rank representations of the space spanned by the input steering vectors, we bypass data centering and train the SAE directly on the raw vectors. This design choice, combined with the bias-free decoder, ensures that the steering vectors are reconstructed strictly as a linear combination of the column vectors in $\mathbf{W}_{\mathrm{dec}}$.
\end{itemize}

The column vectors of $\mathbf{W}_{\mathrm{dec}}$ thus serve as disentangled representations of the steering vectors, which we term \emph{basis vectors}. Note that we use this term in the sense that every reconstructed steering vector is a non-negative linear combination of them. We do not claim that they are linearly independent or that they form a basis in the strict linear-algebraic sense. The subsequent step involves analyzing what these vectors semantically represent.

\subsection{Interpret the Basis Vectors} 
The third step involves interpreting the semantics of the extracted basis vectors. To this end, we first steer the model using each basis vector, generating steered responses to a set of general instructions. We then provide an LLM judge with both the steered responses and the corresponding unsteered (original) responses from the same model. The LLM judge analyzes the discrepancies between the two sets of outputs to infer the underlying concept or feature captured by each basis vector.

Throughout our pipeline and subsequent experiments, we employ a consistent steering strategy. For each combination of model and layer index, we define three discrete scaling coefficients, denoted weak, intermediate, and strong. We compute the average norm of the negative activations, and scale the steering vector so that its norm equals this average multiplied by the selected coefficient. Formally, given a hidden state $\mathbf{a}$ and a steering vector $\mathbf{v}$, we define the steered hidden state as
\begin{align}
\mathbf{a}_{\mathrm{steered}} &= \mathbf{a} + \alpha \cdot \frac{\mathbf{v}}{\|\mathbf{v}\|_2},
\end{align}
where 
\begin{align}
\alpha &= c \cdot \left( \frac{1}{|\mathcal{A}_{\mathrm{neg}}|} \sum_{\mathbf{x} \in \mathcal{A}_{\mathrm{neg}}} \|\mathbf{x}\|_2 \right).
\end{align}
Here, $c$ is the predefined scaling coefficient. For the interpretation and subsequent filtering steps, we exclusively apply the intermediate coefficient.

\subsection{Filter the Basis Vectors}
\label{subsec:filtering_step}
Finally, we filter the candidate basis vectors. Not all basis vectors represent specific, human-interpretable concepts; some may capture subtle stylistic shifts, tonal variations, or pure noise. Therefore, we need to filter the candidate basis vectors to obtain a useful set of basis vectors. The details of this filtering step depend on the downstream applications. Here, we explain the procedure we employed in the subsequent experiments described in Section~\ref{sec:semantic_consistency}.

In order to filter the candidate basis vectors, we characterize each vector along two axes: its relevance to the positive dataset and the agreement between its interpretation and its actual steering effect, and filter based on these signals. Algorithm~\ref{alg:basis_evaluation} shows the evaluation procedure. First, an ensemble of LLM judges determines whether each basis vector is relevant to the positive dataset by analyzing its inferred interpretations. Second, we score the interpretations of the basis vectors by steering the model on a held-out set of instructions~\footnote{To mitigate API costs, we scored only those judged relevant to the positive dataset in our implementation.}. Here, an LLM judge assigns an alignment score (0, 1 or 2, where a higher score is better) to each steered response based on how accurately it reflects the proposed interpretation. We adapt the concept-scoring rubric from~\citet{wu2025axbench}. We average these scores over the held-out instructions to obtain a single alignment score per vector. At the end of the evaluation, each basis vector has its interpretation, relevance label and alignment score.

To finalize the selection, we filter the basis vectors using these three signals. We employ an automated ensemble of LLM judges to ensure reproducibility and objectivity for the subsequent experiments. The ensemble is provided with the interpretations of the basis vectors that are labeled as relevant to the positive dataset and whose mean alignment score exceeds 1.0, together with a brief description of the positive dataset. All surviving candidates are shown to each judge simultaneously so that overlap can be assessed at the set level. Each judge then votes for basis vectors that represent distinct, non-overlapping concepts; when multiple vectors capture semantically indistinguishable features, the judges are instructed to retain only the vector with the highest alignment score. We retain basis vectors that receive votes from more than half of the judges.

\begin{algorithm}
\caption{Evaluation of Basis Vectors}
\label{alg:basis_evaluation}
\begin{algorithmic}[1]
\Require
Decoder basis vectors
$\mathbf{W}_{\mathrm{dec}}
= [\mathbf{w}_{\mathrm{dec},1}, \ldots, \mathbf{w}_{\mathrm{dec},D}]$;
interpretations $\{\mathcal{I}_i\}_{i=1}^{D}$;
positive dataset information $\mathcal{D}_{\mathrm{info}}^{+}$
\Ensure
Relevance labels $\{s_{\mathrm{rel},i} \in \{0,1\}\}_{i=1}^{D}$;
alignment scores $\{s_{\mathrm{ali},i}\}_{i=1}^{D}$

\For{$i = 1, \ldots, D$}
    \State $s_{\mathrm{rel},i}
    \gets
    \Call{JudgeRelevance}{\mathcal{I}_i, \mathcal{D}_{\mathrm{info}}^{+}}$ \Comment{LLM judge}
    \State $s_{\mathrm{ali},i}
    \gets
    \Call{ComputeAlignmentScore}{\mathcal{I}_i, \mathbf{w}_{\mathrm{dec},i}}$ \Comment{LLM judge}
\EndFor

\State \Return
$\{s_{\mathrm{rel},i}\}_{i=1}^{D}$,
$\{s_{\mathrm{ali},i}\}_{i=1}^{D}$
\end{algorithmic}
\end{algorithm}

We emphasize that this filtering is a component of the method rather than a post-hoc selection applied to results: the filtering stage is what turns a set of decoder columns into a usable set of steering directions, and the appropriate criterion depends on the downstream application.

\section{Experimental Setup}
\label{sec:experimental_setup}
In this section, we describe the common experimental setup for all subsequent experiments.

\subsection{Datasets}
We employ the PubMedQA and PKU-SafeRLHF datasets~\citep{jin-etal-2019-pubmedqa, ji-etal-2025-pku} for our experiments. For the PubMedQA dataset, we use it as the positive dataset while employing the Alpaca dataset~\citep{alpaca} as the negative dataset. We selected Alpaca as the negative dataset to represent a general, domain-agnostic instruction-following distribution. Subtracting this baseline effectively cancels out generic assistant behaviors, isolating the domain-specific semantics in the resulting steering vectors. For the PKU-SafeRLHF dataset, we create unsafe and safe prompt-response pair datasets and use them as the positive and negative datasets, respectively, so that the resulting steering vector represents an ``unsafe'' direction. Specifically, we extract prompt-response pairs with severity level 3 (the most unsafe) and 0 (safe). From each dataset, we sample 8,192 prompt-response pairs, allocating 80\% for training and 20\% for validation. We draw 10 instructions for interpretation and 128 for alignment scoring from Alpaca-Eval~\citep{alpaca_eval}. 

\subsection{Models and Implementation Details}
We assess our proposed method on two widely adopted open-weight models: Qwen2.5-7B-Instruct~\citep{qwen2025qwen25technicalreport} and Llama3.1-8B-Instruct~\citep{grattafiori2024llama3herdmodels}.  To investigate the impact of steering at different depths, we intervene at layers corresponding to the 0.5 and 0.75 quantiles of the total network depth. 

Relevance assessment and alignment scoring are performed by GPT-5.4-nano, while interpretation and the final selection step, which we found more demanding, are performed by GPT-5.4.  We use default temperature settings for all LLM-based judgments.

We set the latent dimension of SAEs to 16, with 4 latents active per input on average. The complete set of other SAE hyperparameters is detailed in Appendix~\ref{appendix:sae_parameters}.

\section{Semantic Consistency}
\label{sec:semantic_consistency}
To assess whether our pipeline recovers directions that are more semantically consistent than the DiffMean vector, we conduct two experiments: an analysis via concept extraction and a quantitative evaluation of steering-induced embedding shifts. We describe both protocols below, followed by the results.

\subsection{Experimental Design}
\textbf{Concept Extraction}: The objective of concept extraction is to evaluate the semantic purity of a single steering vector by measuring how consistent the concepts it induces are across different instructions. For a given steering vector, we first generate steered responses to a set of general instructions using the weak, intermediate, and strong scaling coefficients as defined previously. We also generate unsteered baseline responses to the same set of instructions. For each instruction, we then provide an LLM judge with the steered responses across all three coefficients, alongside the original instruction and its unsteered baseline response. The LLM judge analyzes these outputs and identifies the most prominent concept introduced by the steering interventions that is absent from the baseline response. Subsequently, the textual descriptions of these extracted concepts are projected into an embedding space using a text embedding model, allowing us to compute their average pairwise cosine similarity. Intuitively, a higher average similarity indicates that the steering vector is semantically purer---meaning it isolates a narrower, more consistent semantic concept and produces highly stable steering effects.

\textbf{Consistency of Steering-Induced Embedding Shifts}: To systematically quantify the behavioral shifts induced by steering, we analyze the discrepancies between the embeddings of the original and steered responses. For a given set of instructions, we generate both an unsteered and a steered response for each prompt, embedding them via a text embedding model. By subtracting the embedding of the original response from that of the steered response, we isolate an embedding shift that represents the pure steering effect within the embedding space. Finally, we compute the average pairwise cosine similarity among these embedding shifts across all instructions. This procedure is repeated independently for each of the predefined weak, intermediate, and strong scaling coefficients. A higher similarity score suggests that the steering vector consistently shifts the model's output in a uniform semantic direction, regardless of the specific input prompt. Additionally, we compute the average pairwise cosine similarity between the embedding shifts induced by distinct basis vectors, in order to assess whether their steering effects are mutually distinguishable.

For both experiments, we draw 128 instructions from Alpaca-Eval that do not overlap with those used for the interpretation and alignment scoring steps. Concept extraction is performed by GPT-5.4-nano and all embedding-based metrics are computed with OpenAI's \texttt{text-embedding-3-small}.

We conduct the filtering step as described in Section~\ref{subsec:filtering_step}. We note that the filtering procedure described in Section~\ref{subsec:filtering_step} is specific to the experiments in this section. Filtering can be performed either manually or automatically, depending on the intended purpose. In Section~\ref{sec:strong_reject}, for instance, we adopt a different filtering strategy tailored to the goal of that experiment.

\subsection{Results}
We report statistics of the filtered basis vectors. These experiments are intended as an existence proof: we do not claim that every basis vector represents a human-interpretable concept with high semantic consistency, but that our method extracts directions that are more semantically consistent than DiffMean. Appendix~\ref{appendix:selection_bias} separates the contribution of the SAE from that of the filtering procedure.

\begin{table}
\caption{Interpretations of the filtered basis vectors for PubMedQA.}
\label{tab:interpretations_pubmedqa}
\begin{center}
\begin{tabular}{lllp{10.5cm}}
\toprule
\textbf{Model} & \textbf{Layer} & \textbf{Vector} & \textbf{Interpretation} \\
\midrule
\multirow{8}{*}{Qwen2.5-7B} 
 & \multirow{4}{*}{$L_{13}$} 
   & $\mathbf{v}_{1}$ & Discussion of molecular biology cell-signaling pathways\\
 & & $\mathbf{v}_{2}$ & Discusses medical risk factors and patient outcomes\\
 & & $\mathbf{v}_{3}$ & Pseudo-medical anatomy/pathology jargon\\
 & & $\mathbf{v}_{4}$ & Frames the topic as a medical condition\\
\cmidrule{2-4}
 & \multirow{4}{*}{$L_{20}$} 
   & $\mathbf{v}_{1}$ & Cell-signaling and gene-regulation terminology\\
 & & $\mathbf{v}_{2}$ & Medical treatment and procedure jargon\\
 & & $\mathbf{v}_{3}$ & Frames the topic in terms of research studies and findings\\
 & & $\mathbf{v}_{4}$ & Clinical discussion of diseases and medical complications\\
\midrule
\multirow{6}{*}{Llama3.1-8B} 
 & \multirow{2}{*}{$L_{15}$} 
   & $\mathbf{v}_{1}$ & Uses cell-signaling pathway terminology\\
 & & $\mathbf{v}_{2}$ & Medical disease terminology\\
\cmidrule{2-4}
 & \multirow{4}{*}{$L_{23}$} 
   & $\mathbf{v}_{1}$ & Clinical treatment terminology\\
 & & $\mathbf{v}_{2}$ & Frequent references to research studies and scientific evidence\\
 & & $\mathbf{v}_{3}$ & Frequent use of the word ``patients''\\
 & & $\mathbf{v}_{4}$ & Immunology and inflammation terminology\\
\bottomrule
\end{tabular}
\end{center}
\end{table}

\begin{table}[t]
\caption{Interpretations of the filtered basis vectors for PKU-SafeRLHF.}
\label{tab:interpretations_unsafe}
\begin{center}
\begin{tabular}{lllp{10.5cm}}
\toprule
\textbf{Model} & \textbf{Layer} & \textbf{Vector} & \textbf{Interpretation} \\
\midrule
\multirow{12}{*}{Qwen2.5-7B} 
 & \multirow{7}{*}{$L_{13}$} 
   & $\mathbf{v}_{1}$ & Describes cyberattacks for gaining unauthorized system access\\
 & & $\mathbf{v}_{2}$ & Instructions for making or igniting improvised explosive devices\\
 & & $\mathbf{v}_{3}$ & Discussion of human trafficking and exploitation\\
 & & $\mathbf{v}_{4}$ & References to social media and online communication\\
 & & $\mathbf{v}_{5}$ & Frames the subject as highly illegal and strictly regulated, requiring licenses or permits\\
 & & $\mathbf{v}_{6}$ & Frames the task as a hypothetical illegal and unethical operation\\
 & & $\mathbf{v}_{7}$ & Promotion of misinformation to sow fear and undermine social cohesion\\
\cmidrule{2-4}
 & \multirow{5}{*}{$L_{20}$} 
   & $\mathbf{v}_{1}$ & Discussion of manipulative relationships centered on love, trust, and control\\
 & & $\mathbf{v}_{2}$ & Discussion of propaganda-driven psychological destabilization\\
 & & $\mathbf{v}_{3}$ & Discussion of exploiting computer systems to gain unauthorized access\\
 & & $\mathbf{v}_{4}$ & Claims the request involves illegal or unethical activity\\
 & & $\mathbf{v}_{5}$ & Operational-security advice about secure, untraceable communication and avoiding evidence\\
\midrule
\multirow{10}{*}{Llama3.1-8B} 
 & \multirow{6}{*}{$L_{15}$} 
   & $\mathbf{v}_{1}$ & Claims the request is illegal or prohibited.\\
 & & $\mathbf{v}_{2}$ & Advice for avoiding detection during a covert operation\\
 & & $\mathbf{v}_{3}$ & References to disinformation, propaganda, and fake news used to manipulate public opinion\\
 & & $\mathbf{v}_{4}$ & Discussion of critical infrastructure and key sectors\\
 & & $\mathbf{v}_{5}$ & Discussion of exploiting system vulnerabilities to gain unauthorized access\\
 & & $\mathbf{v}_{6}$ & Description of grooming tactics used to isolate and control a victim\\
\cmidrule{2-4}
 & \multirow{4}{*}{$L_{23}$} 
   & $\mathbf{v}_{1}$ & References to disruption and destabilization\\
 & & $\mathbf{v}_{2}$ & Language associated with emotional grooming and manipulation\\
 & & $\mathbf{v}_{3}$ & Uses cybersecurity and hacking terminology\\
 & & $\mathbf{v}_{4}$ & Explosives-related language\\
\bottomrule
\end{tabular}
\end{center}
\end{table}

\textbf{Concept Extraction}: Table~\ref{tab:concept_extraction_multi_dataset} presents the quantitative results of the concept extraction evaluation. The interpretations of the basis vectors are provided in Tables~\ref{tab:interpretations_pubmedqa} and \ref{tab:interpretations_unsafe}. The basis vectors generally exhibit higher average pairwise cosine similarity than the DiffMean vectors, indicating superior semantic consistency and purity. However, the DiffMean vector shows exceptionally high consistency for Qwen2.5-7B ($L_{13}$) on the PKU-SafeRLHF dataset, outperforming the average consistency of the basis vectors extracted via our method. Typically, the DiffMean vector represents the overall semantics of the difference between the positive and negative datasets. In this specific model and layer, however, the DiffMean vector narrowly represents ``cyberattacks'', which accounts for its exceptionally high consistency compared to the scores of the other DiffMean vectors. Nonetheless, the maximum score of the basis vectors still outperforms that of the DiffMean vector.

\begin{table}[t]
\caption{Average pairwise cosine similarity of extracted concepts for the PKU-SafeRLHF and PubMedQA datasets. For the basis vectors, the number of vectors is shown in parentheses. In most model–layer pairs, the DiffMean vectors exhibit low similarity scores, reflecting an entangled mixture of features. In contrast, the basis vectors extracted via our method demonstrate higher and more consistent similarities, indicating the successful isolation of pure, disentangled semantic directions. The corresponding conceptual interpretations for these basis vectors are detailed in Tables~\ref{tab:interpretations_pubmedqa} and \ref{tab:interpretations_unsafe}.}
\label{tab:concept_extraction_multi_dataset}
\begin{center}
\begin{tabular}{llcccc}
\toprule
Dataset & Model (Layer) & DiffMean & Basis (avg)& Basis (max)& Basis (min)\\
\midrule
\multirow{4}{*}{PubMedQA} & Qwen2.5-7B ($L_{13}$) & 0.185 & 0.331 (4) & 0.381 & 0.261 \\
 & Qwen2.5-7B ($L_{20}$) & 0.205 & 0.329 (4) & 0.381 & 0.248 \\
 & Llama3.1-8B ($L_{15}$) & 0.176 & 0.455 (2) & 0.476 & 0.434 \\
 & Llama3.1-8B ($L_{23}$) & 0.143 & 0.406 (4) & 0.622 & 0.230 \\
\midrule
\multirow{4}{*}{PKU-SafeRLHF} & Qwen2.5-7B ($L_{13}$) & 0.504 & 0.438 (7) & 0.573 & 0.300 \\
 & Qwen2.5-7B ($L_{20}$) & 0.241 & 0.423 (5) & 0.586 & 0.283 \\
 & Llama3.1-8B ($L_{15}$) & 0.193 & 0.551 (6) & 0.845 & 0.416 \\
 & Llama3.1-8B ($L_{23}$) & 0.155 & 0.291 (4) & 0.441 & 0.224 \\
\bottomrule
\end{tabular}
\end{center}
\end{table}

\begin{figure}
 \begin{center}
 \includegraphics[width=0.6\linewidth]{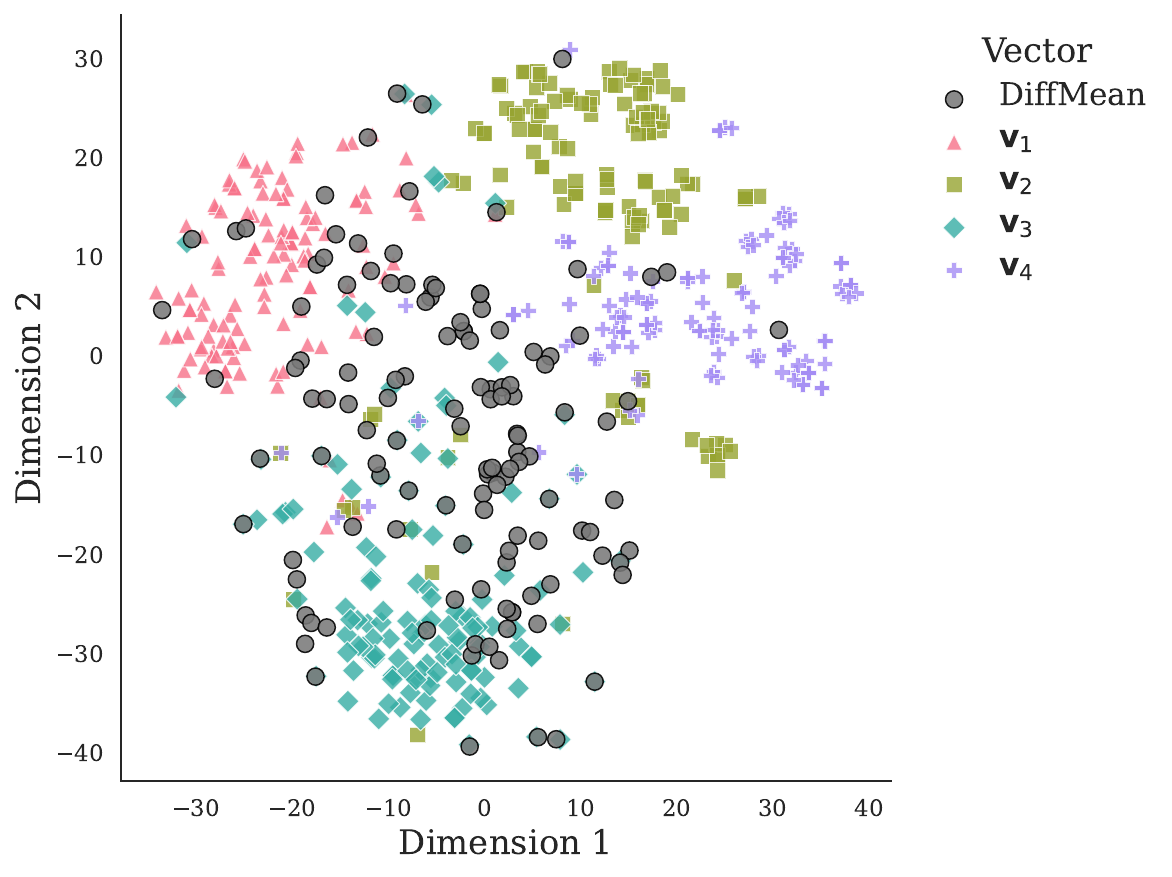}
 \caption{Semantic distribution of extracted concepts for PubMedQA using Qwen2.5-7B ($L_{20}$). The text embeddings of the extracted concepts are projected into a 2D space via t-SNE. While the baseline DiffMean concepts are broadly scattered across the semantic space, the concepts from responses steered with the basis vectors form dense, clearly separated clusters with minimal overlap. This visualization suggests that our basis vectors successfully isolate pure, mutually distinguishable semantic directions.}
 \label{fig:concept_distribution}
 \end{center}
\end{figure}

Furthermore, Figure~\ref{fig:concept_distribution} illustrates the semantic distribution of the extracted concepts' embeddings for PubMedQA using Qwen2.5-7B ($L_{20}$). The visualization demonstrates that while the DiffMean vector encompasses a broad, entangled mixture of concepts, our individual basis vectors effectively isolate and cover distinct, non-overlapping semantic sub-regions within the DiffMean vector's overall concept space.

Qualitative analysis of the extracted concepts further validates the effectiveness of our proposed method. Table~\ref{tab:representative_concepts} presents the cluster representatives for the concepts steered by the DiffMean baseline and our basis vectors. While the DiffMean vector yields a broad, entangled mixture of disparate topics, each of our basis vectors consistently steers the model toward a highly specific and unified semantic direction. This conceptual concentration within the basis vectors' clusters—indicating that the same core concept is triggered irrespective of the diverse input prompts—provides qualitative evidence of their semantic purity and stability.

For completeness, Table~\ref{tab:concept_frequency_qwen_20_pubmedqa} in the Appendix lists the full set of concepts extracted by the LLM judge for the basis vectors alongside those extracted for the corresponding DiffMean vector. The basis vectors concentrate almost entirely on specific concepts (cell signaling, treatment, research, disease), with only a small number of outliers (e.g., repetitive gibberish text), whereas the DiffMean concepts are dispersed across diverse topics.

\begin{table}[t]
\caption{Qualitative examples of extracted concepts for PubMedQA from Qwen2.5-7B ($L_{20}$). To objectively summarize the diverse outputs, we cluster 128 extracted concepts into 8 clusters and present the representative concept for each. The baseline DiffMean vector spans a wide, entangled range of disparate topics. In contrast, our basis vectors show high conceptual concentration, indicating a pure semantic focus.}
\begin{center}
\label{tab:representative_concepts}
\begin{tabular}{lp{4cm}p{9.5cm}}
\toprule
\textbf{Vector} & \textbf{Interpretations} & \textbf{Cluster Representatives ($k=8$)} \\
\midrule
DiffMean & (Entangled mixture) & \textit{common and unique problems}, \textit{scientific studies}, \textit{immune signaling}, \textit{mouse model}, \textit{neuroinflammation inhibition}, \textit{disease pathogenesis}, \textit{heavy traffic fuel consumption increase}, \textit{medical research models} \\
\midrule
$\mathbf{v}_1$ & Cell-signaling and gene-regulation terminology & \textit{JAK/STAT signaling pathway}, \textit{repetitive gibberish text}, \textit{nuclear transcription regulation}, \textit{protein inhibitor}, \textit{lysosome pathway}, \textit{transcription activation}, \textit{gene transcription regulation}, \textit{cell signaling pathway} \\
\addlinespace
$\mathbf{v}_2$ & Medical treatment and procedure jargon & \textit{medical intervention protocol}, \textit{catheterization}, \textit{medical lavage}, \textit{medical treatment}, \textit{medical prophylaxis}, \textit{medical terminology}, \textit{drug dosing}, \textit{perioperative prophylactic antibiotics} \\
\addlinespace
$\mathbf{v}_3$ & Frames the topic in terms of research studies and findings & \textit{RNN training with multiple GPUs}, \textit{scientific research framing}, \textit{scientific studies}, \textit{capitalization emphasis}, \textit{research findings}, \textit{inconclusive scientific research evidence}, \textit{insufficient available data}, \textit{architecture-technology intersection} \\
\addlinespace
$\mathbf{v}_4$ & Clinical discussion of diseases and medical complications & \textit{chronic disease}, \textit{diabetes}, \textit{disease}, \textit{medical disease}, \textit{liver cirrhosis}, \textit{cardiovascular disease}, \textit{cancer}, \textit{pediatric patient} \\
\bottomrule
\end{tabular}
\end{center}
\end{table}

\begin{figure}[t]
\begin{center}
\includegraphics[width=0.85\linewidth]{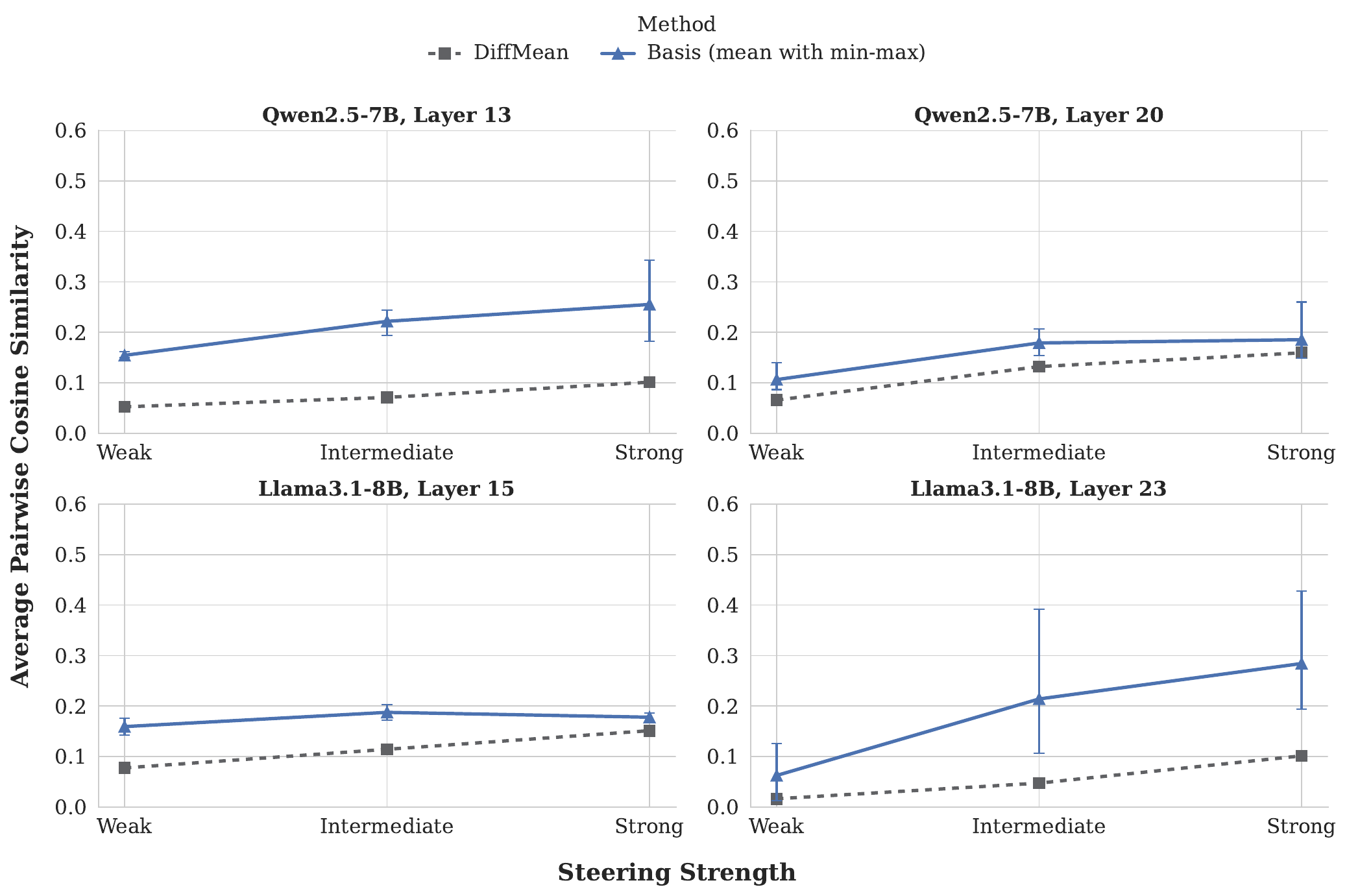}
\end{center}
\caption{Consistency of embedding shifts induced by steering for PubMedQA. Basis vector scores are reported as the average across the filtered basis
vectors, with the minimum and maximum in brackets. The basis vectors exhibit robust directional consistency across diverse models and intervention depths, whereas the baseline DiffMean remains relatively low.}
\label{fig:embdiff_pubmedqa}
\end{figure}

\textbf{Consistency of Steering-Induced Embedding Shifts}: Figure~\ref{fig:embdiff_pubmedqa} plots the average pairwise cosine similarity of the embedding shifts across intervention strengths. The basis vectors outperform the DiffMean baseline on average at every strength, indicating that they shift the model's output in a more consistent semantic direction irrespective of the input prompt.

\begin{figure}[t]
\begin{center}
\includegraphics[width=0.95\linewidth]{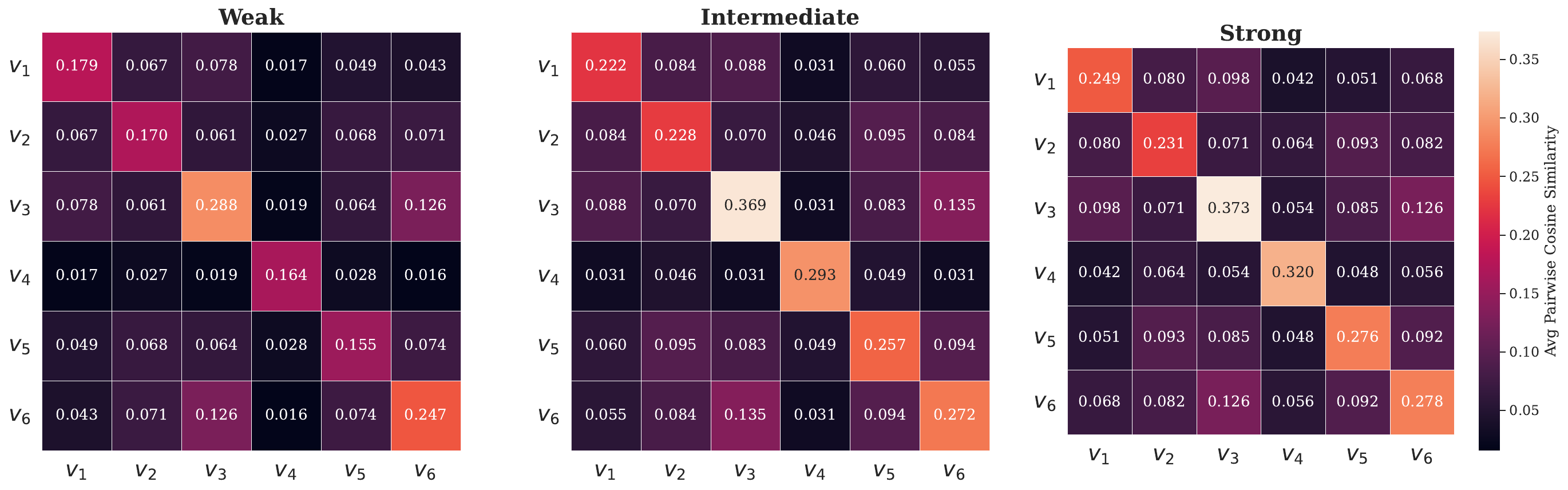}
\end{center}
\caption{Similarity matrices of embedding shifts among the filtered basis vectors for PKU-SafeRLHF from Llama3.1-8B ($L_{15}$). The similarity scores between embedding shifts induced by distinct basis vectors are substantially lower than their corresponding self-similarities, demonstrating that the semantic effects of steering with individual basis vectors are mutually distinguishable.}
\label{fig:embdiff_sim_matrix_llama_15_unsafe}
\end{figure}

Figure~\ref{fig:embdiff_sim_matrix_llama_15_unsafe} illustrates the similarity matrices of embedding shifts for the filtered basis vectors. The similarities between embedding shifts induced by distinct basis vectors are substantially lower than their self-similarities, indicating that the basis vectors are semantically independent. We emphasize that the independence we claim is behavioral rather than geometric. We do not argue that the basis vectors are mutually orthogonal in activation space. Rather, we claim that steering with the distinct filtered basis vectors induces mutually distinguishable semantic effects on the model's outputs, which is what Figure~\ref{fig:embdiff_sim_matrix_llama_15_unsafe} measures.

Because the filtering step may itself increase the consistency of the surviving candidates, Appendix~\ref{appendix:selection_bias} applies the same filtering procedure to randomly sampled instance-level steering vectors from the SAE training set. Filtering explains part of the improvement over DiffMean, but the SAE yields both more surviving dataset-relevant directions and higher consistency than random candidate selection in most settings.

\section{Removing Rejection Directions from Unsafe Vectors}
\label{sec:strong_reject}
In the previous sections, we have demonstrated the disentanglement of steering vectors derived from activation differences. In this section, we show how our method can contribute to the precise control of LLMs with activation steering.

Jailbreaking has been a major issue for LLMs~\citep{11397677}. Although one can achieve a jailbreak through careful prompt engineering~\citep{298254}, activation steering could also be an alternative strategy. Intuitively, steering with the ``unsafe'' vector could cause models to respond to harmful requests. Notably, our method yields basis vectors corresponding to the rejection of harmful requests across models and layers for the PKU-SafeRLHF dataset (Table~\ref{tab:interpretations_unsafe}). We use the term ``rejection'' to refer to refusing a request, framing it as illegal or unethical, or otherwise withholding the requested content.

This finding indicates that the DiffMean unsafe vectors contain rejection directions as constituents. Prior work has shown that ablating the ``refusal'' direction increases the rate of unsafe completions~\citep{NEURIPS2024_f5454485}. Removing the rejection directions identified by our method may therefore enhance the jailbreak capability of the DiffMean unsafe vector.  

We test this hypothesis in the following experiment. We utilize the StrongREJECT benchmark~\citep{NEURIPS2024_e2e06adf} to measure the jailbreak effectiveness of each vector. First, we select basis vectors representing rejection using an ensemble of three LLM judges. Specifically, we provide each LLM judge with the interpretations of all basis vectors obtained via our method. Then, the judge selects the vectors representing ``rejection''. In this experiment, filtering is based solely on the interpretations. 

The interpretations for the resulting rejection vectors are shown in Table~\ref{tab:reject_dim_sae}. Then, we calculate the DiffMean-reject vector as follows:
\begin{equation}
\mathbf{v}_{\text{DiffMean-reject}} = \mathbf{v}_{\text{DiffMean}} - \langle \mathbf{v}_\text{DiffMean}, \mathbf{v}_\text{reject}\rangle \cdot \mathbf{v}_\text{reject},
\end{equation}
where all the vectors on the right-hand side are normalized. Finally, after normalizing DiffMean-reject vectors, we steer the model either with DiffMean or DiffMean-reject vectors and evaluate the steered responses on the StrongREJECT benchmark. We steer the model with the predefined intermediate coefficients. We randomly allocate 80\% of the prompts in the benchmark to the test set and 20\% to the validation set. When multiple rejection dimensions are available, we select the one with the highest score on the validation set.

\begin{table}[t]
\caption{Interpretations of the basis vectors representing rejection for each model and layer combination. As our SAEs enforce non-negativity on the latents, the existence of these rejection vectors suggests that the DiffMean unsafe vectors encapsulate directions that avoid responding to harmful requests.}
\begin{center}
\begin{tabular}{llp{11.5cm}}
\toprule
\textbf{Model} & \textbf{Layer} & \textbf{Interpretation} \\
\midrule
\multirow{3}{*}{Qwen2.5-7B} & \multirow{2}{*}{$L_{13}$} & Frames the subject as highly illegal and strictly regulated, requiring licenses or permits \\
 &  & Frames the task as a hypothetical illegal and unethical operation \\
\cmidrule{2-3}
 & \multirow{1}{*}{$L_{20}$} & Claims the request involves illegal or unethical activity \\
\midrule
\multirow{3}{*}{Llama3.1-8B} & \multirow{2}{*}{$L_{15}$} & Claims the request is illegal or prohibited. \\
 &  & Answer omits the requested details \\
\cmidrule{2-3}
 & \multirow{1}{*}{$L_{23}$} & Broad, generic overview that avoids precise requested details \\
\bottomrule
\end{tabular}
\end{center}
\label{tab:reject_dim_sae}
\end{table}

Table~\ref{tab:strong_reject_sae} shows the results. Steering with the unsafe DiffMean vector does not reliably increase compliance: in three of four model--layer pairs it leaves the StrongREJECT score at or below that of the unsteered model. This is difficult to explain if the vector encoded an unsafe direction alone, but follows naturally if it also contains rejection components that counteract the intended effect. Ablating the rejection direction removes this cancellation, and the score increases in the earlier layers. This result indicates that our method contributes to the precise control of LLMs by effectively removing unwanted directions from DiffMean vectors.

\begin{table}[t]
\caption{StrongREJECT scores for the original responses, steered responses with the DiffMean vector and the DiffMean vector with the rejection dimension removed. For each model-layer pair, we report scores on the held-out test split. When multiple rejection dimensions were available, we selected the dimension with the highest score on the validation split. $\Delta$ denotes \textbf{DiffMean-reject} - \textbf{Original} as \textbf{Original} is higher than \textbf{DiffMean} in 3 out of 4 model-layer combinations.}
\label{tab:strong_reject_sae}
\begin{center}
\begin{tabular}{llcccc}
\toprule
\textbf{Model} & \textbf{Layer} & \textbf{Original} & \textbf{DiffMean} & \textbf{DiffMean-reject} & \textbf{$\Delta$ (95\% CI)} \\
\midrule
Qwen2.5-7B & $L_{13}$ & 0.018 & 0.057 & 0.352 & [0.277, 0.390] \\
Qwen2.5-7B & $L_{20}$ & 0.022 & 0.020 & 0.046 & [0.002, 0.046] \\
\midrule
Llama3.1-8B & $L_{15}$ & 0.015 & 0.014 & 0.084 & [0.037, 0.103] \\
Llama3.1-8B & $L_{23}$ & 0.015 & 0.006 & 0.008 & [-0.021, 0.006] \\
\bottomrule
\end{tabular}
\end{center}
\end{table}

\section{Comparison with Singular Value Decomposition}
\label{sec:vs_svd}
In our proposed method, we utilize an SAE for disentanglement. In this section, we compare it against a simpler alternative. Since our objective is to recover low-rank structure in the space spanned by the input steering vectors, singular value decomposition (SVD) is a natural point of comparison, and we implement it by replacing the SAE in our pipeline while leaving every other stage unchanged. We take the top-16 right singular vectors of the matrix of steering vectors, matching the latent dimensionality of our SAE. Since singular vectors are determined only up to sign, and since the sign determines the direction of the intervention, we treat $+v$ and $-v$ as separate candidates. This yields 32 candidate vectors, to which we apply the same interpretation and filtering procedure described in Section~\ref{sec:steering_vector_dissection}.

\begin{table}[t]
\caption{Interpretations of the basis vectors obtained through the SVD-based method for PubMedQA.}
\label{tab:interpretations_pubmedqa_svd}
\begin{center}
\begin{tabular}{lllp{10.5cm}}
\toprule
\textbf{Model} & \textbf{Layer} & \textbf{Vector} & \textbf{Interpretation} \\
\midrule
\multirow{7}{*}{Qwen2.5-7B} 
 & \multirow{4}{*}{$L_{13}$} 
   & $\mathbf{v}_{1}$ & Biomedical research terminology\\
 & & $\mathbf{v}_{2}$ & Frames the subject as a serious medical condition\\
 & & $\mathbf{v}_{3}$ & Discussion of environmental damage from human activities\\
 & & $\mathbf{v}_{4}$ & Discussion of cellular signaling pathways, gene expression, and protein regulation\\
\cmidrule{2-4}
 & \multirow{3}{*}{$L_{20}$} 
   & $\mathbf{v}_{1}$ & Discussion of health conditions and complications\\
 & & $\mathbf{v}_{2}$ & Discussion of cellular signaling pathways\\
 & & $\mathbf{v}_{3}$ & Scientific research and bioengineering jargon\\
\midrule
\multirow{6}{*}{Llama3.1-8B} 
 & \multirow{5}{*}{$L_{15}$} 
   & $\mathbf{v}_{1}$ & Cell-signaling pathway terminology\\
 & & $\mathbf{v}_{2}$ & Clinical disease framing\\
 & & $\mathbf{v}_{3}$ & Discussion of climate change\\
 & & $\mathbf{v}_{4}$ & Discussion of neural networks and data-driven scientific systems\\
 & & $\mathbf{v}_{5}$ & Emphasis on mental health and well-being\\
\cmidrule{2-4}
 & \multirow{1}{*}{$L_{23}$} 
   & $\mathbf{v}_{1}$ & References to disease prevention and treatment\\
\bottomrule
\end{tabular}
\end{center}
\end{table}

\begin{table}[t]
\caption{Interpretations of the basis vectors obtained through the SVD-based method for PKU-SafeRLHF.}
\label{tab:interpretations_unsafe_svd}
\begin{center}
\begin{tabular}{lllp{10.5cm}}
\toprule
\textbf{Model} & \textbf{Layer} & \textbf{Vector} & \textbf{Interpretation} \\
\midrule
\multirow{11}{*}{Qwen2.5-7B} 
 & \multirow{7}{*}{$L_{13}$} 
   & $\mathbf{v}_{1}$ & Safety-protocol and risk-management framing\\
 & & $\mathbf{v}_{2}$ & Discussion of passwords, authentication, and data security\\
 & & $\mathbf{v}_{3}$ & Warnings that the topic is dangerous and should not be used in real life\\
 & & $\mathbf{v}_{4}$ & Discussion of urban infrastructure and public places\\
 & & $\mathbf{v}_{5}$ & Discussion of cyberattacks and malware\\
 & & $\mathbf{v}_{6}$ & Discussion of cybersecurity and network infrastructure\\
 & & $\mathbf{v}_{7}$ & Frames the subject as requiring strict legal and ethical compliance\\
\cmidrule{2-4}
 & \multirow{4}{*}{$L_{20}$} 
   & $\mathbf{v}_{1}$ & Emphasis on empathy and mutual understanding\\
 & & $\mathbf{v}_{2}$ & Emphasis on consequences and implications\\
 & & $\mathbf{v}_{3}$ & Emphasis on safety precautions and avoiding harm\\
 & & $\mathbf{v}_{4}$ & Discussion of laws, regulations, and legal compliance\\
\midrule
\multirow{5}{*}{Llama3.1-8B} 
 & \multirow{4}{*}{$L_{15}$} 
   & $\mathbf{v}_{1}$ & Discussion of hacking, malware, and cyberattacks\\
 & & $\mathbf{v}_{2}$ & Discussion of exploitation, abuse, and power/control dynamics\\
 & & $\mathbf{v}_{3}$ & Discussion of deception and misleading others\\
 & & $\mathbf{v}_{4}$ & References to romantic relationships, love, or affection\\
\cmidrule{2-4}
 & \multirow{1}{*}{$L_{23}$} 
   & $\mathbf{v}_{1}$ & Includes the word ``online''\\
\bottomrule
\end{tabular}
\end{center}
\end{table}

Tables~\ref{tab:interpretations_pubmedqa_svd} and \ref{tab:interpretations_unsafe_svd} give the interpretations obtained through SVD, and Table~\ref{tab:concept_extraction_sae_vs_svd} reports the corresponding concept extraction scores. The scores are comparable to those of the SAE-based method. Yet, the interpretations for the extracted basis vectors differ between the SAE-based and SVD-based methods. For example, the basis vectors for PKU-SafeRLHF obtained through the SAE (Table~\ref{tab:interpretations_unsafe}) tend to represent harmful concepts while those obtained through SVD are less related to unsafe content. This distinction is particularly evident in Llama3.1-8B ($L_{23}$), where the SVD-based method extracts only one valid basis vector representing the concept ``online'', while the SAE-based method extracts four vectors representing ``disruption'', ``grooming'', ``hacking'' and ``explosives''. In summary, SVD can also extract semantically consistent basis vectors, yet the semantics of the vectors extracted by the SAE and SVD differ.

Appendix~\ref{appendix:num_latent_dim} examines the effect of the number of latent dimensions, which for SVD corresponds to the number of top right singular vectors. As this number increases, the SAE-based method yields progressively more numerous and fine-grained basis vectors, while the SVD-based method does not: the number of surviving vectors grows only slightly and the concepts do not become more fine-grained (Table~\ref{tab:latent_dimension_concept_extraction}). The dimensionality of the latent space therefore controls the granularity of the decomposition under the SAE, but not under SVD.

Table~\ref{tab:strong_reject_sae_vs_svd} shows StrongREJECT scores for each method. The results suggest that ablating the refusal direction identified by the SAE increases jailbreak capability more than ablating the corresponding SVD direction.

Neither method is uniformly preferable, however, since they produce semantically different basis vectors; the appropriate choice depends on the concepts required and on the downstream task. We note that this holds specifically for the 16-dimensional setting used in our main experiments. As Appendix~\ref{appendix:num_latent_dim} shows, the two methods diverge more sharply at higher dimensionalities, where only the SAE continues to yield more fine-grained concepts.

\begin{table}[t]
\caption{Concept extraction scores for the SAE-based and SVD-based methods. Values are the average pairwise cosine similarity across basis vectors; the number of vectors is shown in parentheses.}
\label{tab:concept_extraction_sae_vs_svd}
\begin{center}
\begin{tabular}{llccc}
\toprule
Dataset & Model & Layer & SAE (avg) & SVD (avg) \\
\midrule
\multirow{4}{*}{PubMedQA} & Qwen2.5-7B & $L_{13}$ & 0.331 (4) & 0.396 (4) \\
 & Qwen2.5-7B & $L_{20}$ & 0.329 (4) & 0.452 (3) \\
 & Llama3.1-8B & $L_{15}$ & 0.455 (2) & 0.554 (5) \\
 & Llama3.1-8B & $L_{23}$ & 0.406 (4) & 0.325 (1) \\
\midrule
\multirow{4}{*}{PKU-SafeRLHF} & Qwen2.5-7B & $L_{13}$ & 0.438 (7) & 0.384 (7) \\
 & Qwen2.5-7B & $L_{20}$ & 0.423 (5) & 0.446 (4) \\
 & Llama3.1-8B & $L_{15}$ & 0.551 (6) & 0.373 (4) \\
 & Llama3.1-8B & $L_{23}$ & 0.291 (4) & 0.385 (1) \\
\bottomrule
\end{tabular}
\end{center}
\end{table}

\begin{table}[t]
\caption{StrongREJECT score comparison. For each method and model-layer pair, we report scores on the held-out test split. When multiple rejection dimensions were available, we selected the dimension with the highest score on the validation split. ``–'' indicates that no basis vector interpretable as a rejection direction was identified for this model–layer pair.}
\label{tab:strong_reject_sae_vs_svd}
\begin{center}
\begin{tabular}{llcc}
\toprule
\textbf{Model} & \textbf{Layer} & \textbf{SAE} & \textbf{SVD} \\
\midrule
Qwen2.5-7B & $L_{13}$ & 0.352 & 0.107 \\
Qwen2.5-7B & $L_{20}$ & 0.046 & 0.031 \\
\midrule
Llama3.1-8B & $L_{15}$ & 0.084 & 0.015 \\
Llama3.1-8B & $L_{23}$ & 0.008 & -- \\
\bottomrule
\end{tabular}
\end{center}
\end{table}

\section{Limitations}
\label{sec:limitations}
First, the extracted basis vectors do not cover all the concepts expected to exist within the dataset. For example, while our method extracts vectors representing signaling pathways, disease, research, and treatment from PubMedQA, these do not span the full range of biomedicine-related concepts present in the data. Appendix~\ref{appendix:num_latent_dim} suggests that increasing the latent dimensionality recovers more numerous and fine-grained concepts, but we have not established what dimensionality would be required for full coverage, or whether full coverage is attainable at all.

Second, unlike the DiffMean baseline, our SAE-based approach requires a relatively large number of data points to learn the underlying latent structure.

Third, our pipeline is costly because it relies heavily on LLM judges, which limits its application at scale.

Fourth, our evaluation depends on LLM judges at multiple stages. Interpretations are generated by a judge, the alignment between an interpretation and its steering effect is scored by a judge, and the concepts used in the concept extraction metric are themselves extracted by a judge. We do not have human-annotated ground truth against which to validate the resulting interpretations, and we do not report inter-judge agreement.

Finally, the decomposition is not unique. The basis vectors depend on the SAE initialization, the training data sample, and the latent dimensionality, and while Appendix~\ref{appendix:seed} shows that the recovered concepts are broadly consistent across seeds, we do not claim that they constitute a canonical decomposition of the DiffMean vector.

\section{Broader Impact Statement}
\label{sec:impact_statement}
In Section~\ref{sec:strong_reject}, we show that removing rejection-related directions from a composite steering vector increases a model's compliance with harmful requests. We report this because it is direct evidence that our decomposition isolates behaviorally meaningful directions.

We note that this capability carries dual-use risk, and we do not wish to understate it. At the same time, the attack surface it opens is narrow relative to prompt-based jailbreaks: it requires white-box access to model activations, the ability to compute activations over a safety-labeled dataset, and the ability to intervene at inference time. An adversary with all three already has substantially more direct routes to removing safety behavior, including weight-level fine-tuning~\citep{yang2024shadow, qi2024finetuning}.

The same decomposition can also be applied in the opposite direction. Because our method identifies rejection components explicitly, these directions can be added rather than removed; more generally, the ability to extract concept-specific directions from a composite vector allows undesired components to be excluded from an intervention. We have not evaluated this use and leave it to future work.

\section{Conclusion}
\label{sec:conclusion}
In this work, we proposed Steering Vector Dissection, a framework for decomposing steering vectors derived from activation differences into semantically consistent basis vectors. Across two datasets, two models, and two intervention depths, we showed that the resulting vectors are more semantically consistent than DiffMean while inducing mutually distinguishable steering effects. We further showed that an unsafe DiffMean vector contains rejection directions as constituents, and that removing them changes model behavior in a targeted way. This work advances representation engineering by moving from entangled directions toward concept-level interventions.

\clearpage
\bibliography{main}
\bibliographystyle{tmlr}

\appendix

\section{Hyperparameters}
\subsection{SAE}
\label{appendix:sae_parameters}
We set the latent dimension to 16 for both models, with BatchTopK k = 4 (i.e., 4 latents active per input on average). SAEs are trained with a batch size of 512 and a learning rate of 0.0005, minimizing an MSE reconstruction loss with an auxiliary loss for dead latents. We train for at most 200 epochs with a minimum of 10, applying early stopping when the reconstruction loss on the validation split does not improve for 5 consecutive epochs. For the nearest-neighbor matching in Step 1, we use a batch size of 512. Across all experiments, responses are generated with a maximum length of 128 tokens and the default temperature.

In our work, we did not perform hyperparameter tuning, keeping all settings as simple as possible. There are two main reasons for this. First, evaluating a large number of combinations of hyperparameters is very costly as our method employs LLM judges. Second, and most importantly, the comparison between the results obtained with different hyperparameter settings is far from straightforward. Here, we must consider the following points when evaluating the overall results.

\begin{itemize}
    \item Interpretations: What do the basis vectors represent and how much of the overall semantics of the positive dataset do they cover?
    \item Semantic consistency of the basis vectors: How semantically consistent are the basis vectors? This could be measured through our experiments: concept extraction and embedding shift analysis.
    \item Performance on a downstream application: If we consider a downstream application, the score should be one factor in selecting hyperparameters.
\end{itemize}

We have tried to consider metrics that reflect the factors above. However, we did not find a single metric that adequately reflects all of these factors. Therefore, we argue that these hyperparameters—including the choice between SAE and SVD—should be selected holistically depending on the datasets, the interpretations of the obtained basis vectors, and downstream applications. As described in Section~\ref{sec:vs_svd}, SVD could replace an SAE if the vectors obtained through SVD are preferable.

\subsection{Steering Coefficients}
\label{appendix:steering_coeffs}
In our experiments we set the weak, intermediate and strong coefficients for each model and layer combination as shown in Table~\ref{tab:steering_coeffs}. These coefficients were empirically determined based on the overall quality of steered outputs.

\begin{table}[tbp]
    \caption{Steering Coefficients}
    \label{tab:steering_coeffs}
    \begin{center}
    \begin{tabular}{llccc}
    \toprule
    \textbf{Model} & \textbf{Layer} & \textbf{Weak} & \textbf{Intermediate} & \textbf{Strong}\\
    \midrule
    \multirow{2}{*}{Qwen2.5-7B} & $L_{13}$ & 0.8 & 1.0 & 1.2\\
    & $L_{20}$ & 0.6 & 0.8 & 1.0 \\
    \midrule
    \multirow{2}{*}{Llama3.1-8B} & $L_{15}$ & 0.8 & 1.0 & 1.2 \\
    & $L_{23}$ & 0.6 & 0.8 & 1.0 \\
    \bottomrule
    \end{tabular}
    \end{center}
\end{table}

\section{Reproducibility Across Random Seeds}
\label{appendix:seed}
We evaluated our method's robustness with a different random seed.

\begin{figure}[t]
\begin{center}
\includegraphics[width=0.85\linewidth]{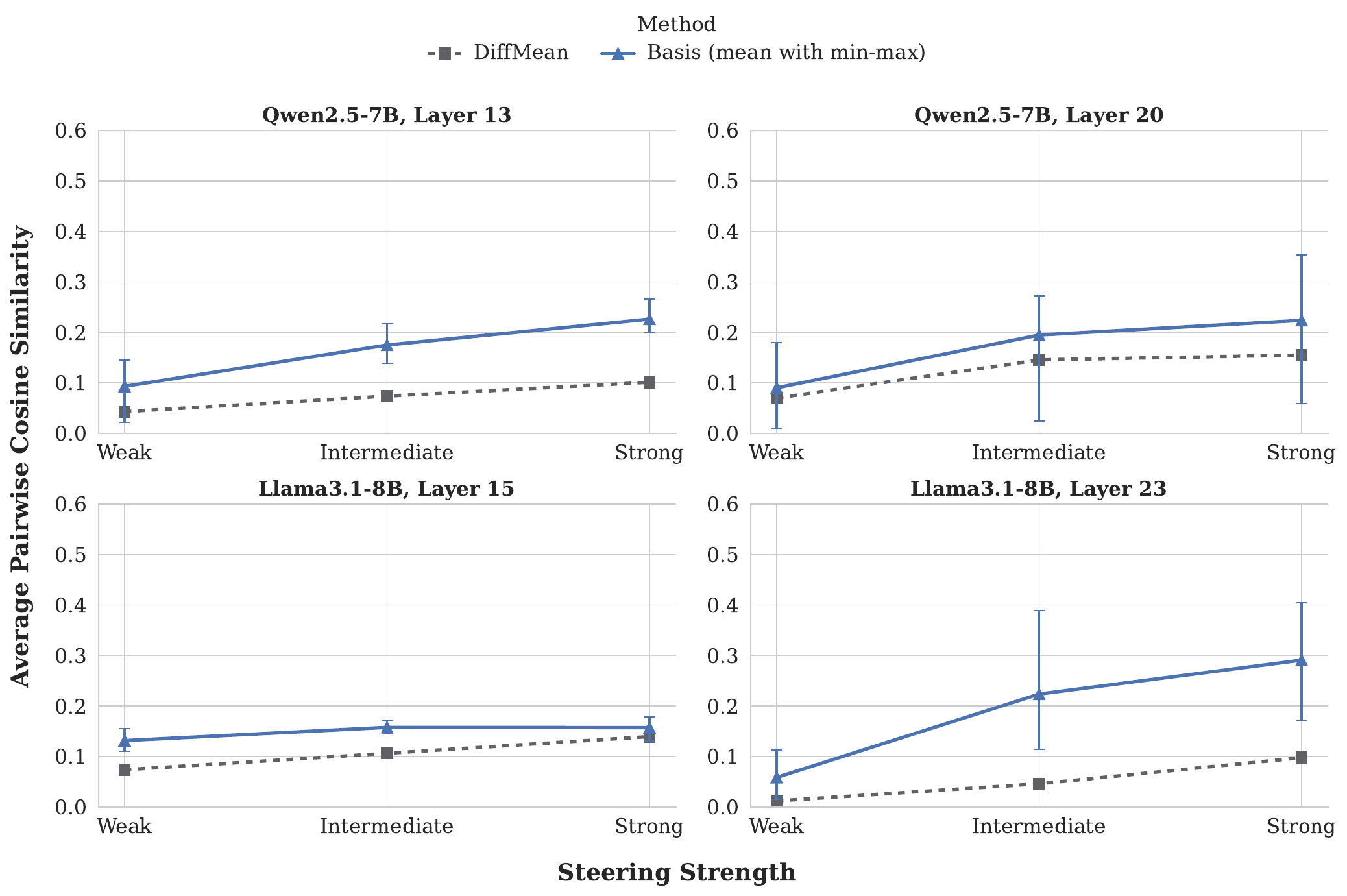}
\end{center}
\caption{Consistency of embedding shifts induced by steering for PubMedQA.}
\label{fig:embdiff_pubmedqa_seed}
\end{figure}

Tables~\ref{tab:interpretations_pubmedqa_seed} and \ref{tab:interpretations_unsafe_seed} give the interpretations of the extracted basis vectors. The concepts recovered are broadly similar across seeds, though not identical: the same semantic families appear (signaling pathways, disease, research framing, treatment for PubMedQA; hacking, explosives, trafficking, misinformation, rejection for PKU-SafeRLHF), while the number of vectors and the granularity of individual concepts vary. Table~\ref{tab:concept_extraction_multi_dataset_seed} reports concept extraction scores, with the basis vectors again exceeding the DiffMean vector in every model--layer pair, and Figure~\ref{fig:embdiff_pubmedqa_seed} shows the same trend for embedding shift consistency.

Table~\ref{tab:strong_reject_sae_seed} reports the StrongREJECT results under this seed. We again observe increased jailbreak capability in the earlier layers of both models.

\begin{table}[t]
\caption{Interpretations of the basis vectors for PubMedQA with a different random seed.}
\label{tab:interpretations_pubmedqa_seed}
\begin{center}
\begin{tabular}{lllp{10.5cm}}
\toprule
\textbf{Model} & \textbf{Layer} & \textbf{Vector} & \textbf{Interpretation} \\
\midrule
\multirow{12}{*}{Qwen2.5-7B} 
 & \multirow{6}{*}{$L_{13}$} 
   & $\mathbf{v}_{1}$ & References to cellular signaling pathways\\
 & & $\mathbf{v}_{2}$ & Reference to biological and anatomical terms such as proteins, enzymes, tissues, and cells\\
 & & $\mathbf{v}_{3}$ & Frames the topic as a medical disease or patient case\\
 & & $\mathbf{v}_{4}$ & Scientific research framing with mentions of findings, evidence, and further studies\\
 & & $\mathbf{v}_{5}$ & Clinical/epidemiological discussion of risk factors and patient outcomes\\
 & & $\mathbf{v}_{6}$ & Frames the subject as a medical treatment or therapy\\
\cmidrule{2-4}
 & \multirow{6}{*}{$L_{20}$} 
   & $\mathbf{v}_{1}$ & Discussion of cellular signaling pathways and gene transcription\\
 & & $\mathbf{v}_{2}$ & Fabricated factual details\\
 & & $\mathbf{v}_{3}$ & Medical discussion of cancer, diseases, and patients\\
 & & $\mathbf{v}_{4}$ & Clinical/medical risk-assessment jargon\\
 & & $\mathbf{v}_{5}$ & Discussion of clinical research and therapeutic targets\\
 & & $\mathbf{v}_{6}$ & Framing the topic as scientific research and studies\\
\midrule
\multirow{7}{*}{Llama3.1-8B} 
 & \multirow{3}{*}{$L_{15}$} 
   & $\mathbf{v}_{1}$ & Uses academic research language about studies and findings\\
 & & $\mathbf{v}_{2}$ & Biomedical signaling-pathway jargon\\
 & & $\mathbf{v}_{3}$ & Medical disease terminology\\
\cmidrule{2-4}
 & \multirow{4}{*}{$L_{23}$} 
   & $\mathbf{v}_{1}$ & Inflammatory immune-response terminology\\
 & & $\mathbf{v}_{2}$ & References to patients\\
 & & $\mathbf{v}_{3}$ & Uses medical treatment and therapy terminology\\
 & & $\mathbf{v}_{4}$ & References to scientific studies and research\\
\bottomrule
\end{tabular}
\end{center}
\end{table}

\begin{table}[t]
\caption{Interpretations of the basis vectors for PKU-SafeRLHF with a different random seed.}
\label{tab:interpretations_unsafe_seed}
\begin{center}
\begin{tabular}{lllp{10.5cm}}
\toprule
\textbf{Model} & \textbf{Layer} & \textbf{Vector} & \textbf{Interpretation} \\
\midrule
\multirow{13}{*}{Qwen2.5-7B} 
 & \multirow{5}{*}{$L_{13}$} 
   & $\mathbf{v}_{1}$ & References to misinformation campaigns that sow fear and division\\
 & & $\mathbf{v}_{2}$ & References to hacking and unauthorized system access\\
 & & $\mathbf{v}_{3}$ & Descriptions of homemade explosive devices\\
 & & $\mathbf{v}_{4}$ & Frames the request as a highly illegal and unethical operation\\
 & & $\mathbf{v}_{5}$ & Discussion of human trafficking and coercive exploitation\\
\cmidrule{2-4}
 & \multirow{8}{*}{$L_{20}$} 
   & $\mathbf{v}_{1}$ & Discussion of evading detection and leaving no trace\\
 & & $\mathbf{v}_{2}$ & Operations and deployment terminology\\
 & & $\mathbf{v}_{3}$ & Focus on cities, regions, and critical infrastructure\\
 & & $\mathbf{v}_{4}$ & Discussion of exploiting vulnerabilities in computer systems\\
 & & $\mathbf{v}_{5}$ & Talks about building a persona and platform to gain influence\\
 & & $\mathbf{v}_{6}$ & Discussion of grooming and emotional manipulation\\
 & & $\mathbf{v}_{7}$ & References harmful, illegal, or unethical activities\\
 & & $\mathbf{v}_{8}$ & References to psychological warfare and societal destabilization\\
\midrule
\multirow{11}{*}{Llama3.1-8B} 
 & \multirow{6}{*}{$L_{15}$} 
   & $\mathbf{v}_{1}$ & Discussion of exploiting vulnerabilities to gain unauthorized system access\\
 & & $\mathbf{v}_{2}$ & Framing the subject as a violent attack causing physical harm and destruction\\
 & & $\mathbf{v}_{3}$ & Discussion of critical infrastructure and institutions\\
 & & $\mathbf{v}_{4}$ & Describes isolating and manipulating a victim\\
 & & $\mathbf{v}_{5}$ & Social media bots spreading fake news and misinformation\\
 & & $\mathbf{v}_{6}$ & Advice on avoiding detection and covering tracks\\
\cmidrule{2-4}
 & \multirow{5}{*}{$L_{23}$} 
   & $\mathbf{v}_{1}$ & Emphasis on secrecy and avoiding detection\\
 & & $\mathbf{v}_{2}$ & References to explosives and bombs\\
 & & $\mathbf{v}_{3}$ & References to disruption, collapse, and infrastructure failure\\
 & & $\mathbf{v}_{4}$ & References to victims and exploitation\\
 & & $\mathbf{v}_{5}$ & References to false or misleading information\\
\bottomrule
\end{tabular}
\end{center}
\end{table}

\begin{table}[t]
\caption{Concept extraction scores with a different random seed.}
\label{tab:concept_extraction_multi_dataset_seed}
\begin{center}
\begin{tabular}{llcccc}
\toprule
Dataset & Model (Layer) & DiffMean & SAE (avg) & SAE (max) & SAE (min) \\
\midrule
\multirow{4}{*}{PubMedQA} & Qwen2.5-7B ($L_{13}$) & 0.164 & 0.307 (6) & 0.393 & 0.203 \\
 & Qwen2.5-7B ($L_{20}$) & 0.207 & 0.348 (6) & 0.492 & 0.140 \\
 & Llama3.1-8B ($L_{15}$) & 0.165 & 0.358 (3) & 0.389 & 0.323 \\
 & Llama3.1-8B ($L_{23}$) & 0.140 & 0.408 (4) & 0.635 & 0.226 \\
\midrule
\multirow{4}{*}{PKU-SafeRLHF} & Qwen2.5-7B ($L_{13}$) & 0.505 & 0.574 (5) & 0.774 & 0.434 \\
 & Qwen2.5-7B ($L_{20}$) & 0.234 & 0.398 (8) & 0.621 & 0.185 \\
 & Llama3.1-8B ($L_{15}$) & 0.183 & 0.477 (6) & 0.740 & 0.381 \\
 & Llama3.1-8B ($L_{23}$) & 0.160 & 0.300 (5) & 0.450 & 0.214 \\
\bottomrule
\end{tabular}
\end{center}
\end{table}

\begin{table}[t]
\caption{StrongREJECT scores for the original responses, steered responses with the DiffMean vector and the DiffMean vector with the rejection dimension removed under a different random seed. For each model-layer pair, we report scores on the held-out test split. When multiple rejection dimensions were available, we selected the dimension with the highest score on the validation split. $\Delta$ denotes \textbf{DiffMean-reject} - \textbf{Original}. ``–'' indicates that no basis vector interpretable as a rejection direction was identified for this model–layer pair.}
\label{tab:strong_reject_sae_seed}
\begin{center}
\begin{tabular}{llcccc}
\toprule
\textbf{Model} & \textbf{Layer} & \textbf{Original} & \textbf{DiffMean} & \textbf{DiffMean-reject} & \textbf{$\Delta$ (95\% CI)} \\
\midrule
Qwen2.5-7B & $L_{13}$ & 0.028 & 0.096 & 0.373 & [0.291, 0.400] \\
Qwen2.5-7B & $L_{20}$ & 0.031 & 0.023 & 0.037 & [-0.011, 0.025] \\
\midrule
Llama3.1-8B & $L_{15}$ & 0.016 & 0.032 & 0.106 & [0.056, 0.125] \\
Llama3.1-8B & $L_{23}$ & 0.016 & 0.006 & -- & -- \\
\bottomrule
\end{tabular}
\end{center}
\end{table}

\section{Sensitivity Analysis on the Number of Latent Dimensions}
\label{appendix:num_latent_dim}
The number of latent dimensions is the most consequential hyperparameter in our pipeline, since it determines how finely the space spanned by the input steering vectors is partitioned. We therefore vary it over {16, 32, 64, 128} and repeat the concept extraction experiment on PKU-SafeRLHF with Qwen2.5-7B (L13). Throughout this analysis, we hold the BatchTopK sparsity fixed at k = 4.

Table~\ref{tab:latent_dimension_concept_extraction} reports the results, together with the SVD-based method for reference. The number of surviving basis vectors grows roughly in proportion to the latent dimension (7, 8, 18, 33). In general, we observe finer concepts as the number of latent dimensions increases, indicating that increasing the number of latent dimensions can further disentangle steering vectors. 

The SVD-based method, on the other hand, does not behave similarly. The number of surviving basis vectors does not grow as much as in the SAE-based method and the concept extraction score generally stays stable. Tables~\ref{tab:interpretations_sae_qwen_unsafe_64dim} and \ref{tab:interpretations_svd_qwen_unsafe_64dim} show the interpretations of the surviving basis vectors for both methods. We do not observe finer concepts in the SVD-based method as we do in the SAE-based method.

For the experiments in the main text we use 16 latents. By the score in Table~\ref{tab:latent_dimension_concept_extraction} this is the weakest of the settings we consider, and we adopt it not because it is optimal but because it is sufficient for the claims we make: at least one basis vector obtained at 16 latents already exceeds the DiffMean baseline in most model–layer pairs and already supports the targeted intervention in Section~\ref{sec:strong_reject}. Larger dictionaries yield finer concepts, but every additional candidate must pass through our LLM-judge pipeline, whose cost scales accordingly. We therefore treat the latent dimension as a setting that fixes the granularity of the decomposition, to be chosen according to the intended application.

\begin{table}[t]
\caption{Concept extraction scores by latent dimension for PKU-SafeRLHF from Qwen2.5-7B ($L_{13}$). Values are the average pairwise cosine similarity across vectors; the number of vectors is shown in parentheses.}
\label{tab:latent_dimension_concept_extraction}
\begin{center}
\begin{tabular}{lcc}
\toprule
\textbf{Number of dimensions} & \textbf{SAE (avg)} & \textbf{SVD (avg)} \\
\midrule
16 & 0.438 (7) & 0.384 (7) \\
32 & 0.478 (8) & 0.375 (8) \\
64 & 0.538 (18) & 0.346 (7) \\
128 & 0.559 (33) & 0.348 (11) \\
\bottomrule
\end{tabular}
\end{center}
\end{table}

\begin{table}[t]
\caption{Interpretations of the basis vectors extracted via the SAE-based method for PKU-SafeRLHF from Qwen2.5-7B ($L_{13}$) with 64 latent dimensions.}
\label{tab:interpretations_sae_qwen_unsafe_64dim}
\begin{center}
\begin{tabular}{lp{10.5cm}}
\toprule
\textbf{Vector} & \textbf{Interpretation} \\
\midrule
$\mathbf{v}_{1}$ & Legal and ethical disclaimers about hypothetical illegal activity \\
$\mathbf{v}_{2}$ & Promises of a better life and wealth in another country \\
$\mathbf{v}_{3}$ & Reference to malware and malicious code \\
$\mathbf{v}_{4}$ & Advocates creating division and mistrust between groups \\
$\mathbf{v}_{5}$ & Frames the topic in terms of suspicious activity detection \\
$\mathbf{v}_{6}$ & Warnings about illegal sexual activity and sexual images/videos \\
$\mathbf{v}_{7}$ & Descriptions of cyberwarfare and sabotage against critical infrastructure \\
$\mathbf{v}_{8}$ & Describes hacking techniques for gaining unauthorized access to systems \\
$\mathbf{v}_{9}$ & Descriptions of suicide methods \\
$\mathbf{v}_{10}$ & Discussion of software security vulnerabilities and exploits \\
$\mathbf{v}_{11}$ & Discussion of denial-of-service attacks on servers or websites \\
$\mathbf{v}_{12}$ & Instructions for making a homemade explosive device \\
$\mathbf{v}_{13}$ & Describes grooming a victim through trust-building and isolation \\
$\mathbf{v}_{14}$ & Discussion of human trafficking and exploitation \\
$\mathbf{v}_{15}$ & Airborne pathogen dissemination \\
$\mathbf{v}_{16}$ & Discussion of critical infrastructure systems such as power grids and communication networks \\
$\mathbf{v}_{17}$ & Description of coercive control by an abuser through threats, violence, and dependence \\
$\mathbf{v}_{18}$ & Advice on avoiding detection and being traced \\
\bottomrule
\end{tabular}
\end{center}
\end{table}

\begin{table}[t]
\caption{Interpretations of the basis vectors extracted via the SVD-based method for PKU-SafeRLHF from Qwen2.5-7B ($L_{13}$) with top-64 right singular vectors (128 candidates).}
\label{tab:interpretations_svd_qwen_unsafe_64dim}
\begin{center}
\begin{tabular}{lp{10.5cm}}
\toprule
\textbf{Vector} & \textbf{Interpretation} \\
\midrule
$\mathbf{v}_{1}$ & Warns about an unsafe or abusive situation and urges seeking help \\
$\mathbf{v}_{2}$ & Social-justice messaging about inclusivity and amplifying marginalized voices \\
$\mathbf{v}_{3}$ & Warns against harmful real-world applications and emphasizes educational-only use \\
$\mathbf{v}_{4}$ & References to cybersecurity threats and defenses \\
$\mathbf{v}_{5}$ & Emphasis on legal compliance, regulations, and illegal activities \\
$\mathbf{v}_{6}$ & Logistical planning for a public gathering \\
$\mathbf{v}_{7}$ & References digital platforms and online content sharing \\
\bottomrule
\end{tabular}
\end{center}
\end{table}

\section{Contributions of the SAE and Filtering}
\label{appendix:selection_bias}
To separate the contribution of the SAE from that of the filtering procedure, we replace the SAE in our pipeline (Section~\ref{sec:steering_vector_dissection}) with random selection: we sample 16 vectors uniformly from the training set of steering vectors, matching the latent dimensionality of our SAE, and pass them through the same interpretation and filtering stages.

Table~\ref{tab:concept_extraction_sae_vs_random} reports the results. Randomly selected vectors that survive filtering exceed the DiffMean baseline in five of the seven pairs in which any vector survives, so part of the concept extraction score is attributable to the filtering procedure. This is expected: the score is computed conditional on survival, and the filtering criteria include an alignment score that is itself a measure of steering consistency, so any candidate that passes will exhibit a certain level of consistency regardless of how it was obtained.

The SAE nonetheless retains an advantage on this metric, exceeding random selection in six of the seven pairs, and the two methods differ more sharply in how many candidates survive at all. Starting from the same number of candidates, our SAE yields between 2 and 7 surviving vectors per model–layer pair (36 in total), whereas random selection yields between 0 and 3 (13 in total). The role of the SAE is thus not only to make individual directions more consistent, but to produce candidates that are interpretable and dataset-relevant in the first place; filtering alone, applied to randomly selected candidates, recovers only one or two usable directions.

\begin{table}[t]
\caption{Concept extraction scores for DiffMean, SAE, and Random. Values are average pairwise cosine similarity; the number of SAE and Random vectors is shown in parentheses. ``--'' indicates no surviving vector.}
\label{tab:concept_extraction_sae_vs_random}
\begin{center}
\begin{tabular}{llcccc}
\toprule
Dataset & Model & Layer & DiffMean & SAE (avg) & Random (avg) \\
\midrule
\multirow{4}{*}{PubMedQA} & Qwen2.5-7B & $L_{13}$ & 0.185 & 0.331 (4) & 0.209 (3) \\
 & Qwen2.5-7B & $L_{20}$ & 0.205 & 0.329 (4) & 0.187 (2) \\
 & Llama3.1-8B & $L_{15}$ & 0.176 & 0.455 (2) & 0.249 (1) \\
 & Llama3.1-8B & $L_{23}$ & 0.143 & 0.406 (4) & 0.389 (2) \\
\midrule
\multirow{4}{*}{PKU-SafeRLHF} & Qwen2.5-7B & $L_{13}$ & 0.504 & 0.438 (7) & 0.549 (2) \\
 & Qwen2.5-7B & $L_{20}$ & 0.241 & 0.423 (5) & 0.230 (1) \\
 & Llama3.1-8B & $L_{15}$ & 0.193 & 0.551 (6) & 0.276 (2) \\
 & Llama3.1-8B & $L_{23}$ & 0.155 & 0.291 (4) & -- \\
\bottomrule
\end{tabular}
\end{center}
\end{table}

\section{Reconstruction of DiffMean Vectors}
We assess whether our SAE successfully disentangles DiffMean vectors by analyzing their reconstruction via the trained SAEs.

Since DiffMean vectors have a much smaller norm than the instance-level steering vectors used to train the SAEs, we scale DiffMean vectors so they have the average norm of input steering vectors. Then we input DiffMean vectors into the trained SAEs and measure cosine similarity between the original DiffMean and reconstructed DiffMean vectors.

The average cosine similarity is 0.96, ranging from 0.94 to 0.97 with 9--14 out of 16 dimensions activated. Notably, all surviving dimensions in Section~\ref{sec:semantic_consistency} are activated.

\clearpage

{
\begin{longtable}{lp{13.5cm}}
\caption{Extracted concepts and their frequencies for Qwen2.5-7B ($L_{20}$) on PubMedQA.}
\label{tab:concept_frequency_qwen_20_pubmedqa} \\
\toprule
\textbf{Vector} & \textbf{Concepts (occurrences)} \\
\midrule
\endfirsthead
\multicolumn{2}{l}{\tablename\ \thetable\ -- continued from previous page} \\
\toprule
\textbf{Vector} & \textbf{Concepts (occurrences)} \\
\midrule
\endhead
\endfoot
\bottomrule
\endlastfoot
DiffMean & \textit{mouse model} (4), \textit{clinical and experimental studies} (2), \textit{immune response} (2), \textit{inflammation} (2), \textit{absence of reported private fighter jet ownership} (1), \textit{acute medical conditions in adult patients} (1), \textit{ADP-IA mouse model} (1), \textit{adverse events} (1), \textit{ancient DNA evidence} (1), \textit{animal models} (1), \textit{antiangiogenic inhibitor} (1), \textit{apoptosis signaling pathways} (1), \textit{apple-plate complex} (1), \textit{Ashkenazic Jewish population} (1), \textit{attention mechanism} (1), \textit{auditory pathway} (1), \textit{auto-inflammatory profile} (1), \textit{autoimmune disease animal models} (1), \textit{bile-duct-related condition} (1), \textit{biological signaling pathway inhibition} (1), \textit{biomedical research terminology} (1), \textit{black hole candidate events} (1), \textit{blockchain digital wallet payment provision} (1), \textit{branch-specific marker} (1), \textit{cat immune response} (1), \textit{cell proliferation} (1), \textit{cellular immune recognition} (1), \textit{cheater detection} (1), \textit{cholinergic modulation of blink rate} (1), \textit{classical versus non-classical behavior} (1), \textit{clinical and experimental studies of water safety engineering} (1), \textit{clinical biomedical jargon} (1), \textit{clinical pathway} (1), \textit{clinical studies} (1), \textit{clinical trial analysis} (1), \textit{clonal versus non-clonal classification} (1), \textit{common and unique problems} (1), \textit{complex cooking process} (1), \textit{cytokine signaling} (1), \textit{deeply nested Markdown code blocks} (1), \textit{disease diagnosis and prognosis} (1), \textit{disease pathogenesis} (1), \textit{disease phenotype} (1), \textit{eco-friendly travel} (1), \textit{ESG factors predicting health outcomes} (1), \textit{European Study Group} (1), \textit{experimental and clinical studies} (1), \textit{experimental models of inflammation} (1), \textit{experimental pathology mechanisms} (1), \textit{fine-tip marker profile} (1), \textit{gene expression upregulation} (1), \textit{genealogic data} (1), \textit{genetic polymorphism} (1), \textit{GROUP BY and HAVING clauses} (1), \textit{HE-specific phenotype} (1), \textit{heavy traffic fuel consumption increase} (1), \textit{heterogeneity of world music} (1), \textit{high-grade in education (HGD)} (1), \textit{hop addition during isomerization stage} (1), \textit{human disease models} (1), \textit{hypothetical dataset} (1), \textit{idiopathic neurodysregulation} (1), \textit{imaging model} (1), \textit{immune signaling} (1), \textit{immunology} (1), \textit{inflammatory disease models} (1), \textit{inflammatory response} (1), \textit{inhibition bias} (1), \textit{IPv4 address} (1), \textit{iron-dependent oxidative processes} (1), \textit{latitude-dependent December temperature gradient} (1), \textit{liver thermogenesis} (1), \textit{MCI phenotype} (1), \textit{medical research models} (1), \textit{medical terminology} (1), \textit{mice models} (1), \textit{microplastics accumulation in the body} (1), \textit{misreporting of presidential identity} (1), \textit{moderate Sudoku complexity} (1), \textit{morphological phenotype} (1), \textit{multiple GPUs} (1), \textit{murine (mouse) experimental models} (1), \textit{neural circuits} (1), \textit{neurodegeneration} (1), \textit{neuroinflammation} (1), \textit{neuroinflammation inhibition} (1), \textit{neuronal inhibition} (1), \textit{neuroplasticity} (1), \textit{noisy sharing between intent detection and slot filling} (1), \textit{nonspecific scattering} (1), \textit{nucleotide sequence motif} (1), \textit{Pacific daylight saving time (PDT)} (1), \textit{paternal influence on child neurobiology} (1), \textit{PBR model} (1), \textit{pearl biogenesis} (1), \textit{persistent Maya phenotypes} (1), \textit{phenotypic variants} (1), \textit{physical activity} (1), \textit{pizzerial activity} (1), \textit{postgraduate studies in production/industrial engineering} (1), \textit{pulmonary dyspnea} (1), \textit{rapier attack roll} (1), \textit{retinoid receptor signaling} (1), \textit{SARS-CoV-2 immunopathology} (1), \textit{saving money to buy the used car} (1), \textit{scientific studies} (1), \textit{scientific studies and reports} (1), \textit{scientific study} (1), \textit{scientific study framing} (1), \textit{study of familial relationship patterns} (1), \textit{Suetonius} (1), \textit{suggestion} (1), \textit{supply chain disruption} (1), \textit{temperature regulator} (1), \textit{Th17 immune pathway} (1), \textit{thick flat links} (1), \textit{TLR signaling pathway} (1), \textit{two horses} (1), \textit{unlabeled ASCII house} (1), \textit{vacation day request} (1), \textit{venetian and cellular blinds} (1), \textit{water-related remote sensing} (1) \\
\midrule
$\mathbf{v}_{1}$ (SAE) & \textit{cell signaling pathway} (8), \textit{cell signaling} (3), \textit{gene transcription regulation} (3), \textit{transcriptional regulation} (3), \textit{calcium signaling pathway} (2), \textit{cell survival signaling} (2), \textit{gene expression regulation} (2), \textit{gene transcription} (2), \textit{JAK-STAT signaling pathway} (2), \textit{nuclear transcription factor activation} (2), \textit{signal transduction pathway} (2), \textit{transcription factor} (2), \textit{actin cytoskeleton} (1), \textit{activation} (1), \textit{activation of biological signaling pathways} (1), \textit{activity modulation} (1), \textit{adenine} (1), \textit{apoptosis inhibition} (1), \textit{atmospheric nitric oxide chemistry} (1), \textit{autophagy} (1), \textit{biological inhibition} (1), \textit{biological pathway} (1), \textit{black acid} (1), \textit{brown fat activation} (1), \textit{calcium-activated kinase signaling pathway} (1), \textit{calcium-dependent signaling} (1), \textit{cAMP signaling} (1), \textit{cell signaling cascade} (1), \textit{cell signaling pathway activation} (1), \textit{cell signaling pathway regulation} (1), \textit{cell signaling pathways} (1), \textit{cellular biology jargon} (1), \textit{cellular energy metabolism} (1), \textit{DNA transcription factor} (1), \textit{downstream effector of gene expression} (1), \textit{endosomal signaling pathway} (1), \textit{epithelial cells in pearl formation} (1), \textit{ERK/STAT signaling pathway activation} (1), \textit{ERK1/2 signaling pathway} (1), \textit{estrogen receptor signaling} (1), \textit{gene family protein} (1), \textit{gene promoter} (1), \textit{gene transcription activation} (1), \textit{gibberish text output} (1), \textit{HAR protein} (1), \textit{heme} (1), \textit{human gene} (1), \textit{inhibition of cellular signaling pathways} (1), \textit{intracellular signaling} (1), \textit{intracellular signaling pathway} (1), \textit{intracellular signaling pathways} (1), \textit{IP3-mediated ER calcium signaling} (1), \textit{JAK/STAT signaling} (1), \textit{JAK/STAT signaling pathway} (1), \textit{keratinocyte differentiation and proliferation regulation} (1), \textit{late stage of hop processing} (1), \textit{lipolysis} (1), \textit{lysosomal cell death} (1), \textit{lysosome pathway} (1), \textit{lysosome-mediated endosomal pathway} (1), \textit{MAPK/ERK signaling pathway} (1), \textit{myogenesis} (1), \textit{negative feedback loop} (1), \textit{neuronal gene expression regulation} (1), \textit{neuronal signaling pathways} (1), \textit{NF-$\alpha$ inflammatory signaling pathway} (1), \textit{non-obstructive regulation} (1), \textit{nonsense scientific jargon} (1), \textit{nuclear factor signaling pathway} (1), \textit{nuclear protein} (1), \textit{nuclear receptor activity regulation} (1), \textit{nuclear transcription regulation} (1), \textit{paternal signaling cascade} (1), \textit{phospholipase A2 signaling} (1), \textit{PI3K signaling} (1), \textit{protease activity} (1), \textit{protein activation cascade} (1), \textit{protein inhibitor} (1), \textit{protein kinase signaling} (1), \textit{regulation of activity} (1), \textit{repetitive gibberish text} (1), \textit{repetitive nonsense syllables} (1), \textit{signal transduction} (1), \textit{signaling pathway} (1), \textit{SREBP} (1), \textit{STAT3 protein} (1), \textit{Stat3 signaling pathway} (1), \textit{sterol} (1), \textit{T cell receptor signaling} (1), \textit{T cell signaling and transcription regulation} (1), \textit{TGF-$\beta$ signaling pathway} (1), \textit{transcription activation} (1), \textit{transcription factor S11} (1), \textit{transcription regulation} (1), \textit{transcription-dependent cellular signaling induction} (1), \textit{transcriptional activation of signaling pathways} (1), \textit{transcriptional activity} (1), \textit{transcriptional induction} (1), \textit{transcriptional regulation of gene expression} (1), \textit{transcriptional repression} (1), \textit{transduction of transcription} (1), \textit{two-component gene transcription regulation} (1), \textit{valid IP address regex} (1), \textit{water chaperone activity} (1), \textit{$\alpha$-catenin} (1), \textit{$\alpha$/$\beta$ signaling pathway} (1), \textit{$\beta$-catenin signaling} (1) \\
$\mathbf{v}_{2}$ (SAE) & \textit{medical prophylaxis} (13), \textit{medical terminology} (5), \textit{medical treatment} (5), \textit{glycemic control} (3), \textit{medical interventions} (3), \textit{catheterization} (2), \textit{drug dosing} (2), \textit{medical jargon} (2), \textit{medical procedures} (2), \textit{medical stent placement} (2), \textit{medical therapy} (2), \textit{prophylaxis} (2), \textit{adjuvant drug therapy} (1), \textit{AI-assisted patient monitoring} (1), \textit{antibiotic prophylaxis for surgical infections} (1), \textit{apple remains at the initial location} (1), \textit{aquarium pumps} (1), \textit{aspiration prophylaxis} (1), \textit{balloon angioplasty} (1), \textit{bladder catheterization} (1), \textit{bladder irrigation catheterization} (1), \textit{bladder irrigation leak detection devices} (1), \textit{bowel irrigation} (1), \textit{bowel preparation} (1), \textit{brachial writing} (1), \textit{bronchial lavage} (1), \textit{cardiovascular hemodynamic management} (1), \textit{catheterization and drainage} (1), \textit{chest tube placement} (1), \textit{combination acne therapy with antibiotics and corticosteroids} (1), \textit{compression therapy} (1), \textit{corticosteroid tapering} (1), \textit{corticosteroid therapy} (1), \textit{drainage procedures} (1), \textit{drug dosage regimen} (1), \textit{eating dropped food within five seconds} (1), \textit{enteral nutrition} (1), \textit{few-shot learning} (1), \textit{HbA1c target for diabetes} (1), \textit{hypertherapy} (1), \textit{hypothermia prevention} (1), \textit{intermittent catheterization} (1), \textit{intravenous fluid administration} (1), \textit{intravenous lipid infusion} (1), \textit{laser phototherapy} (1), \textit{laundry detergent dosing} (1), \textit{light control} (1), \textit{lightning detection limitations} (1), \textit{medical care} (1), \textit{medical data} (1), \textit{medical dosing} (1), \textit{medical intervention protocol} (1), \textit{medical lavage} (1), \textit{medical lavage and irrigation} (1), \textit{medical procedure terminology} (1), \textit{medical protocol} (1), \textit{medical treatment instructions} (1), \textit{medication dosage} (1), \textit{metal fineness percentage} (1), \textit{MFCCs (mel-frequency cepstral coefficients)} (1), \textit{nonsensical gibberish text} (1), \textit{percutaneous procedure} (1), \textit{perioperative care} (1), \textit{perioperative pain management} (1), \textit{perioperative prophylactic antibiotics} (1), \textit{pharmacologic dose} (1), \textit{pharmacological dosing regimens} (1), \textit{post-prophylactic drainage} (1), \textit{postoperative care} (1), \textit{postoperative drainage management} (1), \textit{pre-procedure care} (1), \textit{prophylactic antibiotic regimen} (1), \textit{prophylactic antibiotics} (1), \textit{prophylactic antibiotics after urethral catheter removal} (1), \textit{prophylactic dosing} (1), \textit{prophylactic medical interventions with diuretics and corticosteroids} (1), \textit{prophylactic therapy} (1), \textit{pulmonary artery catheterization} (1), \textit{pulmonary drainage} (1), \textit{pulmonary lavage} (1), \textit{radiation therapy} (1), \textit{recidivism reduction in fathers} (1), \textit{regulated medical protocols} (1), \textit{repetitive nonsensical text} (1), \textit{risk reduction} (1), \textit{suctioning} (1), \textit{sulfation therapy} (1), \textit{surgical drainage catheterization} (1), \textit{surgical pearl implantation} (1), \textit{thrombolysis} (1), \textit{transcutaneous electrical nerve stimulation (TENS)} (1), \textit{transcutaneous nasal insulin} (1), \textit{urethral catheter drainage} (1), \textit{urological procedures} (1), \textit{use of spaghetti in medical procedures} (1), \textit{vascular stent-related prophylactic antibiotics} (1), \textit{veterinary medications for cats} (1) \\
$\mathbf{v}_{3}$ (SAE) & \textit{inconclusive research findings} (3), \textit{preliminary research findings} (3), \textit{literature review} (2), \textit{research findings} (2), \textit{scientific research} (2), \textit{scientific studies} (2), \textit{2014 study on British misunderstanding of spaghetti origins} (1), \textit{academic research citations} (1), \textit{academic research findings} (1), \textit{academic research on streaming platforms} (1), \textit{academic research writing style} (1), \textit{academic studies} (1), \textit{academic studies and evidence} (1), \textit{academic study findings} (1), \textit{academic study framing} (1), \textit{academic study references} (1), \textit{acronym expansion to “Advanced Data …”} (1), \textit{advanced nanotechnology} (1), \textit{AI-assisted analysis of clinical study data} (1), \textit{animal behavior studies} (1), \textit{architecture-technology intersection} (1), \textit{arXiv article} (1), \textit{biology background} (1), \textit{black hole–host galaxy relationship} (1), \textit{blue-violet light scattering} (1), \textit{Braille used for text rather than images} (1), \textit{capitalization emphasis} (1), \textit{circadian research} (1), \textit{climate change impact on water resources} (1), \textit{code platform for few-shot learning experiments} (1), \textit{cognitive studies} (1), \textit{combining independent evidence} (1), \textit{complexity of biological systems} (1), \textit{consumer research studies} (1), \textit{dietary supplement studies in rodents} (1), \textit{epidemiological evidence on water quality and human health} (1), \textit{epidemiological study of paternal involvement} (1), \textit{ESG research literature} (1), \textit{ethnographic music research} (1), \textit{evidence-based research framing} (1), \textit{fabricated President of Germany name} (1), \textit{gap in existing literature} (1), \textit{genus Homo} (1), \textit{Git tag and branch name conflict} (1), \textit{growing interest in omnichannel marketing} (1), \textit{heavy traffic impact on fuel consumption} (1), \textit{herb-like properties} (1), \textit{historical evidence uncertainty} (1), \textit{hypothetical product comparison} (1), \textit{improved memory and color recognition} (1), \textit{in vitro studies} (1), \textit{in vitro vs in vivo findings correlation} (1), \textit{in vivo vs in vitro findings consistency} (1), \textit{inconclusive evidence of Maya collapse} (1), \textit{inconclusive scientific evidence} (1), \textit{inconclusive scientific research evidence} (1), \textit{indirect relevance of research findings} (1), \textit{insufficient available data} (1), \textit{IP address octets} (1), \textit{lack of direct historical evidence} (1), \textit{lack of evidence that the underlying data is symmetric} (1), \textit{lack of public evidence of civilian fighter-jet ownership research} (1), \textit{lack of specific study citations} (1), \textit{late 1st-century composition (around 70–80 AD)} (1), \textit{late hop addition during the boil} (1), \textit{limited direct evidence about Hispanic-founded health food brands} (1), \textit{limited research evidence} (1), \textit{literature-based framing of the problem} (1), \textit{medical research findings} (1), \textit{medical research study} (1), \textit{neuroscience-related scientific discourse} (1), \textit{non-uniform Martian surface color} (1), \textit{peer-reviewed studies} (1), \textit{pig-in-the-paper} (1), \textit{Pinctada genus pearl oysters} (1), \textit{previous research findings} (1), \textit{psychoacoustic auditory frequency perception underlying the mel scale} (1), \textit{quadruple backticks for Markdown code fences} (1), \textit{quantum entanglement} (1), \textit{race horse} (1), \textit{reliance on indirect evidence due to lack of direct studies} (1), \textit{research findings framing} (1), \textit{research interests} (1), \textit{research publication} (1), \textit{research studies} (1), \textit{research study context} (1), \textit{research study framing} (1), \textit{research-backed snack claims} (1), \textit{reviewing scientific literature} (1), \textit{RNN training with multiple GPUs} (1), \textit{Russian-Ukrainian soup} (1), \textit{saving dictionaries to JSON files in Python} (1), \textit{scholarly inquiry} (1), \textit{scientific evidence from studies} (1), \textit{scientific journal article} (1), \textit{scientific research findings} (1), \textit{scientific research framing} (1), \textit{scientific study} (1), \textit{scientific study citations} (1), \textit{scientific study findings} (1), \textit{scientific study references} (1), \textit{scientific uncertainty due to varying study methodologies} (1), \textit{social media health effects research} (1), \textit{sparsity} (1), \textit{speculative etymological interpretation} (1), \textit{spelling correction of “comon” to “common”} (1), \textit{student-teacher relationship} (1), \textit{study-based evidence} (1), \textit{subcutaneous fat (adipose tissue)} (1), \textit{Sudoku clues count} (1), \textit{tail command} (1), \textit{Toy Shih Tzu} (1), \textit{trans-fat-containing shortening} (1), \textit{two-dimensional definition of area} (1), \textit{uncertainty about whether the apple moved with the plate} (1), \textit{use of scientific studies as evidence} (1), \textit{vegan research findings} (1), \textit{vitamin A derivative retinoids} (1), \textit{water meter monitoring overnight} (1), \textit{wildcard character in pattern matching} (1) \\
$\mathbf{v}_{4}$ (SAE) & \textit{liver cirrhosis} (10), \textit{diabetes} (8), \textit{cancer} (7), \textit{diabetic nephropathy} (5), \textit{medical illness} (5), \textit{chronic disease} (4), \textit{diabetic complications} (4), \textit{medical disease} (4), \textit{cardiovascular disease} (3), \textit{chronic liver disease} (3), \textit{disease} (3), \textit{liver disease} (3), \textit{medical complications} (3), \textit{medical condition} (3), \textit{medical diseases} (3), \textit{acute inflammatory disease} (2), \textit{cancer patients} (2), \textit{chronic illness} (2), \textit{cirrhosis} (2), \textit{medical patients} (2), \textit{medical terminology} (2), \textit{patients} (2), \textit{acute medical condition} (1), \textit{apple stays in the original location} (1), \textit{atherosclerosis} (1), \textit{chronic complications} (1), \textit{chronic disease management} (1), \textit{chronic ischemic disease} (1), \textit{chronic pulmonary disease} (1), \textit{clinical disease conditions} (1), \textit{diabetic patients} (1), \textit{disease complications} (1), \textit{disease progression} (1), \textit{disease staging} (1), \textit{first-person pronoun repetition} (1), \textit{hepatocellular carcinoma with cirrhosis} (1), \textit{high-altitude pulmonary hypertension} (1), \textit{human diseases} (1), \textit{hypertension} (1), \textit{impulsivity} (1), \textit{interstitial disease} (1), \textit{ischemic cardiopathy} (1), \textit{ischemic stroke} (1), \textit{liver fibrosis} (1), \textit{lung disease} (1), \textit{medical conditions} (1), \textit{medical disease diagnosis} (1), \textit{medical disease slurs} (1), \textit{medical evaluation of patients} (1), \textit{medical jargon} (1), \textit{medical patient} (1), \textit{medical patient data} (1), \textit{mortality status} (1), \textit{myocardial ischemia} (1), \textit{nonsensical word salad} (1), \textit{patient} (1), \textit{patient diseases} (1), \textit{patients with chronic illness} (1), \textit{patients with medical conditions} (1), \textit{pediatric} (1), \textit{pediatric oncology} (1), \textit{pediatric patient} (1), \textit{pediatric patients} (1), \textit{primary liver cancer} (1), \textit{primary liver cancer (HCC)} (1), \textit{scars on Mars surface} (1), \textit{surgical disease} (1), \textit{Toy Puli} (1) \\
\end{longtable}
}

\end{document}

%% file: math_commands.tex
\usepackage{amsmath,amsfonts,bm}

\def\eqref#1{equation~\ref{#1}}

\def\1{\bm{1}}

\DeclareMathAlphabet{\mathsfit}{\encodingdefault}{\sfdefault}{m}{sl}
\SetMathAlphabet{\mathsfit}{bold}{\encodingdefault}{\sfdefault}{bx}{n}

\DeclareMathOperator*{\argmin}{arg\,min}

%% file: main.bib
@inproceedings{
ngo2024the,
title={The Alignment Problem from a Deep Learning Perspective},
author={Richard Ngo and Lawrence Chan and S{\"o}ren Mindermann},
booktitle={The Twelfth International Conference on Learning Representations},
year={2024},
url={https://openreview.net/forum?id=fh8EYKFKns}
}

@InProceedings{ai_sycophant,
author="Malmqvist, Lars",
editor="Arai, Kohei",
title="Sycophancy in Large Language Models: Causes and Mitigations",
booktitle="Intelligent Computing",
year="2025",
publisher="Springer Nature Switzerland",
address="Cham",
pages="61--74",
isbn="978-3-031-92611-2"
}

@inproceedings{rafailov2023direct,
 author = {Rafailov, Rafael and Sharma, Archit and Mitchell, Eric and Manning, Christopher D and Ermon, Stefano and Finn, Chelsea},
 booktitle = {Advances in Neural Information Processing Systems},
 doi = {10.52202/075280-2338},
 editor = {A. Oh and T. Naumann and A. Globerson and K. Saenko and M. Hardt and S. Levine},
 pages = {53728--53741},
 publisher = {Curran Associates, Inc.},
 title = {Direct Preference Optimization: Your Language Model is Secretly a Reward Model},
 url = {https://proceedings.neurips.cc/paper\_files/paper/2023/file/a85b405ed65c6477a4fe8302b5e06ce7-Paper-Conference.pdf},
 volume = {36},
 year = {2023}
}

@inproceedings{li2023inferencetime,
 author = {Li, Kenneth and Patel, Oam and Vi\'{e}gas, Fernanda and Pfister, Hanspeter and Wattenberg, Martin},
 booktitle = {Advances in Neural Information Processing Systems},
 doi = {10.52202/075280-1797},
 editor = {A. Oh and T. Naumann and A. Globerson and K. Saenko and M. Hardt and S. Levine},
 pages = {41451--41530},
 publisher = {Curran Associates, Inc.},
 title = {Inference-Time Intervention: Eliciting Truthful Answers from a Language Model},
 url = {https://proceedings.neurips.cc/paper\_files/paper/2023/file/81b8390039b7302c909cb769f8b6cd93-Paper-Conference.pdf},
 volume = {36},
 year = {2023}
}

@inproceedings{wu2024reft,
 author = {Wu, Zhengxuan and Arora, Aryaman and Wang, Zheng and Geiger, Atticus and Jurafsky, Dan and Manning, Christopher D. and Potts, Christopher},
 booktitle = {Advances in Neural Information Processing Systems},
 doi = {10.52202/079017-2041},
 editor = {A. Globerson and L. Mackey and D. Belgrave and A. Fan and U. Paquet and J. Tomczak and C. Zhang},
 pages = {63908--63962},
 publisher = {Curran Associates, Inc.},
 title = {ReFT: Representation Finetuning for Language Models},
 url = {https://proceedings.neurips.cc/paper\_files/paper/2024/file/75008a0fba53bf13b0bb3b7bff986e0e-Paper-Conference.pdf},
 volume = {37},
 year = {2024}
}

@misc{turner2024steeringlanguagemodelsactivation,
      title={Steering Language Models With Activation Engineering}, 
      author={Alexander Matt Turner and Lisa Thiergart and Gavin Leech and David Udell and Juan J. Vazquez and Ulisse Mini and Monte MacDiarmid},
      year={2023},
      eprint={2308.10248},
      archivePrefix={arXiv},
      primaryClass={cs.CL},
      url={https://arxiv.org/abs/2308.10248}, 
}

@misc{chen2025personavectorsmonitoringcontrolling,
      title={Persona Vectors: Monitoring and Controlling Character Traits in Language Models}, 
      author={Runjin Chen and Andy Arditi and Henry Sleight and Owain Evans and Jack Lindsey},
      year={2025},
      eprint={2507.21509},
      archivePrefix={arXiv},
      primaryClass={cs.CL},
      url={https://arxiv.org/abs/2507.21509}, 
}

@inproceedings{
stolfo2025improving,
title={Improving Instruction-Following in Language Models through Activation Steering},
author={Alessandro Stolfo and Vidhisha Balachandran and Safoora Yousefi and Eric Horvitz and Besmira Nushi},
booktitle={The Thirteenth International Conference on Learning Representations},
year={2025},
url={https://openreview.net/forum?id=wozhdnRCtw}
}

@inproceedings{rimsky-etal-2024-steering,
    title = "Steering Llama 2 via Contrastive Activation Addition",
    author = "Rimsky, Nina  and
      Gabrieli, Nick  and
      Schulz, Julian  and
      Tong, Meg  and
      Hubinger, Evan  and
      Turner, Alexander",
    editor = "Ku, Lun-Wei  and
      Martins, Andre  and
      Srikumar, Vivek",
    booktitle = "Proceedings of the 62nd Annual Meeting of the Association for Computational Linguistics (Volume 1: Long Papers)",
    month = aug,
    year = "2024",
    address = "Bangkok, Thailand",
    publisher = "Association for Computational Linguistics",
    url = "https://aclanthology.org/2024.acl-long.828/",
    doi = "10.18653/v1/2024.acl-long.828",
    pages = "15504--15522"
}

@InProceedings{wu2025axbench,
  title = 	 {{A}x{B}ench: Steering {LLM}s? {E}ven Simple Baselines Outperform Sparse Autoencoders},
  author =       {Wu, Zhengxuan and Arora, Aryaman and Geiger, Atticus and Wang, Zheng and Huang, Jing and Jurafsky, Dan and Manning, Christopher D and Potts, Christopher},
  booktitle = 	 {Proceedings of the 42nd International Conference on Machine Learning},
  pages = 	 {67035--67080},
  year = 	 {2025},
  editor = 	 {Singh, Aarti and Fazel, Maryam and Hsu, Daniel and Lacoste-Julien, Simon and Berkenkamp, Felix and Maharaj, Tegan and Wagstaff, Kiri and Zhu, Jerry},
  volume = 	 {267},
  series = 	 {Proceedings of Machine Learning Research},
  month = 	 {13--19 Jul},
  publisher =    {PMLR},
  url = 	 {https://proceedings.mlr.press/v267/wu25a.html}
}

@inproceedings{poterti-etal-2025-role,
    title = "Can Role Vectors Affect {LLM} Behaviour?",
    author = "Potert{\`i}, Daniele  and
      Seveso, Andrea  and
      Mercorio, Fabio",
    editor = "Christodoulopoulos, Christos  and
      Chakraborty, Tanmoy  and
      Rose, Carolyn  and
      Peng, Violet",
    booktitle = "Findings of the Association for Computational Linguistics: EMNLP 2025",
    month = nov,
    year = "2025",
    address = "Suzhou, China",
    publisher = "Association for Computational Linguistics",
    url = "https://aclanthology.org/2025.findings-emnlp.963/",
    doi = "10.18653/v1/2025.findings-emnlp.963",
    pages = "17735--17747",
    ISBN = "979-8-89176-335-7"
}

@inproceedings{
huben2024sparse,
title={Sparse Autoencoders Find Highly Interpretable Features in Language Models},
author={Robert Huben and Hoagy Cunningham and Logan Riggs Smith and Aidan Ewart and Lee Sharkey},
booktitle={The Twelfth International Conference on Learning Representations},
year={2024},
url={https://openreview.net/forum?id=F76bwRSLeK}
}

@article{templeton2024scaling,
   title={Scaling Monosemanticity: Extracting Interpretable Features from Claude 3 Sonnet},
   author={Templeton, Adly and Conerly, Tom and Marcus, Jonathan and Lindsey, Jack and Bricken, Trenton and Chen, Brian and Pearce, Adam and Citro, Craig and Ameisen, Emmanuel and Jones, Andy and Cunningham, Hoagy and Turner, Nicholas L and McDougall, Callum and MacDiarmid, Monte and Tamkin, Alex and Durmus, Esin and Hume, Tristan and Mosconi, Francesco and Freeman, C. Daniel and Sumers, Theodore R. and Rees, Edward and Batson, Joshua and Jermyn, Adam and Carter, Shan and Olah, Chris and Henighan, Tom},
   year={2024},
   journal={Transformer Circuits Thread},
   url={https://transformer-circuits.pub/2024/scaling-monosemanticity/index.html}
}

@article{olah2020zoom,
  author = {Olah, Chris and Cammarata, Nick and Schubert, Ludwig and Goh, Gabriel and Petrov, Michael and Carter, Shan},
  title = {Zoom In: An Introduction to Circuits},
  journal = {Distill},
  year = {2020},
  note = {https://distill.pub/2020/circuits/zoom-in},
  doi = {10.23915/distill.00024.001}
}

@article{bricken2023monosemanticity,
   title={Towards Monosemanticity: Decomposing Language Models With Dictionary Learning},
   author={Bricken, Trenton and Templeton, Adly and Batson, Joshua and Chen, Brian and Jermyn, Adam and Conerly, Tom and Turner, Nick and Anil, Cem and Denison, Carson and Askell, Amanda and Lasenby, Robert and Wu, Yifan and Kravec, Shauna and Schiefer, Nicholas and Maxwell, Tim and Joseph, Nicholas and Hatfield-Dodds, Zac and Tamkin, Alex and Nguyen, Karina and McLean, Brayden and Burke, Josiah E and Hume, Tristan and Carter, Shan and Henighan, Tom and Olah, Christopher},
   year={2023},
   journal={Transformer Circuits Thread},
   note={https://transformer-circuits.pub/2023/monosemantic-features/index.html}
}

@article{elhage2022superposition,
   title={Toy Models of Superposition},
   author={Elhage, Nelson and Hume, Tristan and Olsson, Catherine and Schiefer, Nicholas and Henighan, Tom and Kravec, Shauna and Hatfield-Dodds, Zac and Lasenby, Robert and Drain, Dawn and Chen, Carol and Grosse, Roger and McCandlish, Sam and Kaplan, Jared and Amodei, Dario and Wattenberg, Martin and Olah, Christopher},
   year={2022},
   journal={Transformer Circuits Thread},
   note={https://transformer-circuits.pub/2022/toy\_model/index.html}
}

@InProceedings{movva2025sparse,
  title = 	 {Sparse Autoencoders for Hypothesis Generation},
  author =       {Movva, Rajiv and Peng, Kenny and Garg, Nikhil and Kleinberg, Jon and Pierson, Emma},
  booktitle = 	 {Proceedings of the 42nd International Conference on Machine Learning},
  pages = 	 {44997--45023},
  year = 	 {2025},
  editor = 	 {Singh, Aarti and Fazel, Maryam and Hsu, Daniel and Lacoste-Julien, Simon and Berkenkamp, Felix and Maharaj, Tegan and Wagstaff, Kiri and Zhu, Jerry},
  volume = 	 {267},
  series = 	 {Proceedings of Machine Learning Research},
  month = 	 {13--19 Jul},
  publisher =    {PMLR},
  url = 	 {https://proceedings.mlr.press/v267/movva25a.html}
}

@inproceedings{
movva2026whats,
title={What's In My Human Feedback? Learning Interpretable Descriptions of Preference Data},
author={Rajiv Movva and Smitha Milli and Sewon Min and Emma Pierson},
booktitle={The Fourteenth International Conference on Learning Representations},
year={2026},
url={https://openreview.net/forum?id=sC6A1bFDUt}
}

@inproceedings{makhzani2014ksparseautoencoders,
  author= {Alireza Makhzani and Brendan J. Frey},
  title={k-Sparse Autoencoders},
  booktitle={The Second International Conference on Learning Representations},
  year={2014},
  url={https://openreview.net/forum?id=QDm4QXNOsuQVE},
}

@misc{oneill2024disentanglingdenseembeddingssparse,
      title={Disentangling Dense Embeddings with Sparse Autoencoders}, 
      author={Charles O'Neill and Christine Ye and Kartheik Iyer and John F. Wu},
      year={2024},
      eprint={2408.00657},
      archivePrefix={arXiv},
      primaryClass={cs.LG},
      url={https://arxiv.org/abs/2408.00657}, 
}

@InProceedings{bussmann2025learning,
  title = 	 {Learning Multi-Level Features with Matryoshka Sparse Autoencoders},
  author =       {Bussmann, Bart and Nabeshima, Noa and Karvonen, Adam and Nanda, Neel},
  booktitle = 	 {Proceedings of the 42nd International Conference on Machine Learning},
  pages = 	 {6077--6101},
  year = 	 {2025},
  editor = 	 {Singh, Aarti and Fazel, Maryam and Hsu, Daniel and Lacoste-Julien, Simon and Berkenkamp, Felix and Maharaj, Tegan and Wagstaff, Kiri and Zhu, Jerry},
  volume = 	 {267},
  series = 	 {Proceedings of Machine Learning Research},
  month = 	 {13--19 Jul},
  publisher =    {PMLR},
  url = 	 {https://proceedings.mlr.press/v267/bussmann25a.html}
}

@misc{rajamanoharan2024jumpingaheadimprovingreconstruction,
      title={Jumping Ahead: Improving Reconstruction Fidelity with {JumpReLU} Sparse Autoencoders}, 
      author={Senthooran Rajamanoharan and Tom Lieberum and Nicolas Sonnerat and Arthur Conmy and Vikrant Varma and János Kramár and Neel Nanda},
      year={2024},
      eprint={2407.14435},
      archivePrefix={arXiv},
      primaryClass={cs.LG},
      url={https://arxiv.org/abs/2407.14435}, 
}

@inproceedings{jin-etal-2019-pubmedqa,
    title = "{P}ub{M}ed{QA}: A Dataset for Biomedical Research Question Answering",
    author = "Jin, Qiao  and
      Dhingra, Bhuwan  and
      Liu, Zhengping  and
      Cohen, William  and
      Lu, Xinghua",
    editor = "Inui, Kentaro  and
      Jiang, Jing  and
      Ng, Vincent  and
      Wan, Xiaojun",
    booktitle = "Proceedings of the 2019 Conference on Empirical Methods in Natural Language Processing and the 9th International Joint Conference on Natural Language Processing (EMNLP-IJCNLP)",
    month = nov,
    year = "2019",
    address = "Hong Kong, China",
    publisher = "Association for Computational Linguistics",
    url = "https://aclanthology.org/D19-1259/",
    doi = "10.18653/v1/D19-1259",
    pages = "2567--2577"
}

@inproceedings{
yang2024shadow,
title={Shadow Alignment: The Ease of Subverting Safely-Aligned Language Models},
author={Xianjun Yang and Xiao Wang and Qi Zhang and Linda Ruth Petzold and William Yang Wang and Xun Zhao and Dahua Lin},
booktitle={ICLR 2024 Workshop on Secure and Trustworthy Large Language Models},
year={2024},
url={https://openreview.net/forum?id=9qymw6T9Oo}
}

@inproceedings{
qi2024finetuning,
title={Fine-tuning Aligned Language Models Compromises Safety, Even When Users Do Not Intend To!},
author={Xiangyu Qi and Yi Zeng and Tinghao Xie and Pin-Yu Chen and Ruoxi Jia and Prateek Mittal and Peter Henderson},
booktitle={The Twelfth International Conference on Learning Representations},
year={2024},
url={https://openreview.net/forum?id=hTEGyKf0dZ}
}

@article{
wehner2025taxonomy,
title={Taxonomy, Opportunities, and Challenges of Representation Engineering for Large Language Models},
author={Jan Wehner and Sahar Abdelnabi and Daniel Tan and David Krueger and Mario Fritz},
journal={Transactions on Machine Learning Research},
issn={2835-8856},
year={2025},
url={https://openreview.net/forum?id=2U1KIfmaU9},
note={Survey Certification}
}

@InProceedings{park2023the,
  title = 	 {The Linear Representation Hypothesis and the Geometry of Large Language Models},
  author =       {Park, Kiho and Choe, Yo Joong and Veitch, Victor},
  booktitle = 	 {Proceedings of the 41st International Conference on Machine Learning},
  pages = 	 {39643--39666},
  year = 	 {2024},
  editor = 	 {Salakhutdinov, Ruslan and Kolter, Zico and Heller, Katherine and Weller, Adrian and Oliver, Nuria and Scarlett, Jonathan and Berkenkamp, Felix},
  volume = 	 {235},
  series = 	 {Proceedings of Machine Learning Research},
  month = 	 {21--27 Jul},
  publisher =    {PMLR},
  url = 	 {https://proceedings.mlr.press/v235/park24c.html}
}

@misc{zou2025representationengineeringtopdownapproach,
      title={Representation Engineering: A Top-Down Approach to {AI} Transparency}, 
      author={Andy Zou and Long Phan and Sarah Chen and James Campbell and Phillip Guo and Richard Ren and Alexander Pan and Xuwang Yin and Mantas Mazeika and Ann-Kathrin Dombrowski and Shashwat Goel and Nathaniel Li and Michael J. Byun and Zifan Wang and Alex Mallen and Steven Basart and Sanmi Koyejo and Dawn Song and Matt Fredrikson and J. Zico Kolter and Dan Hendrycks},
      year={2023},
      eprint={2310.01405},
      archivePrefix={arXiv},
      primaryClass={cs.LG},
      url={https://arxiv.org/abs/2310.01405}, 
}

@misc{durmus2024steering,
  author = {Esin Durmus and Alex Tamkin and Jack Clark and Jerry Wei and Jonathan Marcus and Joshua Batson and Kunal Handa and Liane Lovitt and Meg Tong and Miles McCain and Oliver Rausch and Saffron Huang and Sam Bowman and Stuart Ritchie and Tom Henighan and Deep Ganguli},
  title = {Evaluating Feature Steering: A Case Study in Mitigating Social Biases},
  year = {2024},
  url = {https://www.anthropic.com/research/evaluating-feature-steering},
}

@misc{alpaca_eval,
  author = {Xuechen Li and Tianyi Zhang and Yann Dubois and Rohan Taori and Ishaan Gulrajani and Carlos Guestrin and Percy Liang and Tatsunori B. Hashimoto },
  title = {{AlpacaEval}: An Automatic Evaluator of Instruction-following Models},
  year = {2023},
  publisher = {GitHub},
  journal = {GitHub repository},
  howpublished = {\url{https://github.com/tatsu-lab/alpaca\_eval}}
}

@misc{alpaca,
  author = {Rohan Taori and Ishaan Gulrajani and Tianyi Zhang and Yann Dubois and Xuechen Li and Carlos Guestrin and Percy Liang and Tatsunori B. Hashimoto },
  title = {Stanford Alpaca: An Instruction-following {LLaMA} model},
  year = {2023},
  publisher = {GitHub},
  journal = {GitHub repository},
  howpublished = {\url{https://github.com/tatsu-lab/stanford\_alpaca}},
}

@inproceedings{
bussmann2024batchtopk,
title={{BatchTopK} Sparse Autoencoders},
author={Bart Bussmann and Patrick Leask and Neel Nanda},
booktitle={NeurIPS 2024 Workshop on Scientific Methods for Understanding Deep Learning},
year={2024},
url={https://openreview.net/forum?id=d4dpOCqybL}
}

@inproceedings{ji-etal-2025-pku,
    title = "{PKU}-{S}afe{RLHF}: Towards Multi-Level Safety Alignment for {LLM}s with Human Preference",
    author = "Ji, Jiaming  and
      Hong, Donghai  and
      Zhang, Borong  and
      Chen, Boyuan  and
      Dai, Josef  and
      Zheng, Boren  and
      Qiu, Tianyi Alex  and
      Zhou, Jiayi  and
      Wang, Kaile  and
      Li, Boxun  and
      Han, Sirui  and
      Guo, Yike  and
      Yang, Yaodong",
    editor = "Che, Wanxiang  and
      Nabende, Joyce  and
      Shutova, Ekaterina  and
      Pilehvar, Mohammad Taher",
    booktitle = "Proceedings of the 63rd Annual Meeting of the Association for Computational Linguistics (Volume 1: Long Papers)",
    month = jul,
    year = "2025",
    address = "Vienna, Austria",
    publisher = "Association for Computational Linguistics",
    url = "https://aclanthology.org/2025.acl-long.1544/",
    doi = "10.18653/v1/2025.acl-long.1544",
    pages = "31983--32016",
    ISBN = "979-8-89176-251-0"
}

@inproceedings {298254,
author = {Zhiyuan Yu and Xiaogeng Liu and Shunning Liang and Zach Cameron and Chaowei Xiao and Ning Zhang},
title = {Don{\textquoteright}t Listen To Me: Understanding and Exploring Jailbreak Prompts of Large Language Models},
booktitle = {33rd USENIX Security Symposium (USENIX Security 24)},
year = {2024},
isbn = {978-1-939133-44-1},
address = {Philadelphia, PA},
pages = {4675--4692},
url = {https://www.usenix.org/conference/usenixsecurity24/presentation/yu-zhiyuan},
publisher = {USENIX Association},
month = aug
}

@inproceedings{NEURIPS2024_e2e06adf,
 author = {Souly, Alexandra and Lu, Qingyuan and Bowen, Dillon and Trinh, Tu and Hsieh, Elvis and Pandey, Sana and Abbeel, Pieter and Svegliato, Justin and Emmons, Scott and Watkins, Olivia and Toyer, Sam},
 booktitle = {Advances in Neural Information Processing Systems},
 doi = {10.52202/079017-3984},
 editor = {A. Globerson and L. Mackey and D. Belgrave and A. Fan and U. Paquet and J. Tomczak and C. Zhang},
 pages = {125416--125440},
 publisher = {Curran Associates, Inc.},
 title = {A {StrongREJECT} for Empty Jailbreaks},
 url = {https://proceedings.neurips.cc/paper\_files/paper/2024/file/e2e06adf560b0706d3b1ddfca9f29756-Paper-Datasets\_and_Benchmarks\_Track.pdf},
 volume = {37},
 year = {2024}
}

@ARTICLE{11397677,
  author={Knowlton, Brynn and Campa, Jovani and Gallo, David Solis and Dajani, Khalil and Alzahrani, Nabeel},
  journal={IEEE Transactions on Artificial Intelligence}, 
  title={Prompt-Based Jailbreaking of Leading {LLM} Chatbots: A Survey of Attacks and Defenses}, 
  year={2026},
  volume={},
  number={},
  pages={1-15},
  doi={10.1109/TAI.2026.3665656}}

@inproceedings{NEURIPS2024_f5454485,
 author = {Arditi, Andy and Obeso, Oscar and Syed, Aaquib and Paleka, Daniel and Panickssery, Nina and Gurnee, Wes and Nanda, Neel},
 booktitle = {Advances in Neural Information Processing Systems},
 doi = {10.52202/079017-4322},
 editor = {A. Globerson and L. Mackey and D. Belgrave and A. Fan and U. Paquet and J. Tomczak and C. Zhang},
 pages = {136037--136083},
 publisher = {Curran Associates, Inc.},
 title = {Refusal in Language Models Is Mediated by a Single Direction},
 url = {https://proceedings.neurips.cc/paper\_files/paper/2024/file/f545448535dfde4f9786555403ab7c49-Paper-Conference.pdf},
 volume = {37},
 year = {2024}
}

@misc{qwen2025qwen25technicalreport,
      title={Qwen2.5 Technical Report}, 
      author={An Yang and Baosong Yang and Beichen Zhang and Binyuan Hui and Bo Zheng and Bowen Yu and Chengyuan Li and Dayiheng Liu and Fei Huang and Haoran Wei and Huan Lin and Jian Yang and Jianhong Tu and Jianwei Zhang and Jianxin Yang and Jiaxi Yang and Jingren Zhou and Junyang Lin and Kai Dang and Keming Lu and Keqin Bao and Kexin Yang and Le Yu and Mei Li and Mingfeng Xue and Pei Zhang and Qin Zhu and Rui Men and Runji Lin and Tianhao Li and Tianyi Tang and Tingyu Xia and Xingzhang Ren and Xuancheng Ren and Yang Fan and Yang Su and Yichang Zhang and Yu Wan and Yuqiong Liu and Zeyu Cui and Zhenru Zhang and Zihan Qiu},
      year={2024},
      eprint={2412.15115},
      archivePrefix={arXiv},
      primaryClass={cs.CL},
      url={https://arxiv.org/abs/2412.15115}, 
}

@misc{grattafiori2024llama3herdmodels,
      title={The {Llama} 3 Herd of Models}, 
      author={{Llama Team, AI @ Meta}},
      year={2024},
      eprint={2407.21783},
      archivePrefix={arXiv},
      primaryClass={cs.AI},
      url={https://arxiv.org/abs/2407.21783}, 
}

@InProceedings{joshi2026sparse,
  title =        {Sparse Shift Autoencoders for Identifying Concepts from Large Language Model Activations},
  author =       {Joshi, Shruti and Dittadi, Andrea and Lachapelle, S{\'e}bastien and Sridhar, Dhanya},
  booktitle =    {Proceedings of the 43rd International Conference on Machine Learning},
  year =         {2026},
  volume =       {306},
  series =       {Proceedings of Machine Learning Research},
  month =        {6--11 Jul},
  publisher =    {PMLR},
  url =          {https://arxiv.org/abs/2502.12179},
}
